\documentclass[runningheads]{llncs}

\usepackage{eccv}

\usepackage{eccvabbrv}
\usepackage{marvosym}

\usepackage{graphicx}
\usepackage{booktabs}
\usepackage{colortbl}
\usepackage{wrapfig}

\usepackage[accsupp]{axessibility}

\usepackage{hyperref}

\usepackage{orcidlink}

\newif\ifmanualspacing

\usepackage{amsmath,amsfonts,bm}
\usepackage{xcolor}

\usepackage{pifont}

\definecolor{RankFirst}{RGB}{190,0,0}
\definecolor{RankSecond}{RGB}{0,90,180}
\definecolor{badbad}{RGB}{180,180,180}
\newcommand{\rankfirst}[1]{\textcolor{RankFirst}{#1}}
\newcommand{\ranksecond}[1]{\textcolor{RankSecond}{#1}}
\newcommand{\Bad}[1]{\textcolor{badbad}{#1}}

\def\tabref#1{Tab.~\ref{#1}}

\def\figref#1{Fig.~\ref{#1}}

\def\eqref#1{(\ref{#1})}

\def\1{\bm{1}}

\DeclareMathAlphabet{\mathsfit}{\encodingdefault}{\sfdefault}{m}{sl}
\SetMathAlphabet{\mathsfit}{bold}{\encodingdefault}{\sfdefault}{bx}{n}

\usepackage{graphicx}
\usepackage{url}
\usepackage{booktabs}
\usepackage{nicefrac}
\usepackage{microtype}
\usepackage{wrapfig}

\usepackage{enumitem}

\usepackage[ruled,noend,linesnumbered]{algorithm2e}

\usepackage{multirow}
\usepackage{tablefootnote}

\usepackage{color}
\usepackage{colortbl}
\usepackage{xcolor}

\definecolor{blgrey}{rgb}{0.6,0.6,0.6}
\definecolor{bblue}{rgb}{0.855,0.933,0.98}
\definecolor{dblue}{HTML}{5297D6}
\definecolor{gainred}{rgb}{0.1,0.5,0.3}
\definecolor{citecolor}{HTML}{0071BC}
\definecolor{linkcolor}{HTML}{ED1C24}

\usepackage{listings}
\usepackage{color}

\definecolor{dkcyan}{cmyk}{1,0,0,.25}
\definecolor{dkgreen}{rgb}{0,0.6,0}
\definecolor{gray}{rgb}{0.5,0.5,0.5}
\definecolor{mauve}{rgb}{0.58,0,0.82}

\begin{document}

\title{Motion-aware Sparse Pipeline for Lightweight Object Tracking}

\author{Qingmao Wei\inst{1,2}\orcidlink{0000-0001-9982-3119} \and
Fagui Liu\inst{1,2}\orcidlink{0000-0003-1135-4982}\textsuperscript{\Letter} \and
Dengke Zhang\inst{1}\orcidlink{0009-0001-2941-0084} \and
Qingze He\inst{1}\orcidlink{0009-0005-0859-2770} \and
Quan Tang\inst{2}\orcidlink{0000-0003-4011-6166}\textsuperscript{\Letter}
}

\authorrunning{Q.~Wei et al.}

\institute{South China University of Technology, Guangzhou, China \and
Pengcheng Laboratory, Shenzhen, China}

\maketitle
\begingroup
\renewcommand{\thefootnote}{\Letter}
\footnotetext{Corresponding authors: Quan Tang (\email{tangq@pcl.ac.cn}) and Fagui Liu (\email{fgliu@scut.edu.cn}).}
\endgroup

\begin{abstract}
  Transformer-based object trackers are renowned for their strong performance, yet dense token processing often leads to prohibitive computational cost, limiting real-time deployment on edge devices. While recent works explore token pruning to reduce computation, they often stop short of an end-to-end sparse pipeline, as early-layer token scores can be noisy without a motion prior, and many trackers ultimately fall back to dense reshaping  to feed the dense prediction head that partially negates the savings.
  We introduce Motion-aware Sparse Tracker (MaST), a sparse tracking framework that makes sparsity effective from tokens to boxes. First, MaST injects a lightweight motion prior to refine cross-attention-based importance scores, enabling earlier and more stable token reduction in the search region. Second, we introduce a natively sparse prediction head that operates directly on the retained unstructured tokens with a score-first, regress-once design, eliminating dense padding/reshaping and reducing redundant computation.
  Extensive experiments on multiple benchmarks demonstrate that MaST establishes new state of the art among lightweight trackers, where MaST-tiny attains 63.8 AUC on LaSOT and 80.1 SUC on TrackingNet, surpassing the prior best AsymTrack-S by +1.0 AUC and +2.2 SUC 
  while running at 152 FPS on Jetson Nano, nearly twice as fast as AsymTrack-S at 88 FPS.  Code is available at \url{github.com/TsingWei/MaST}.
\end{abstract}

\section{Introduction}

\begin{figure*}[t]
  \centering
  \includegraphics[width=\linewidth]{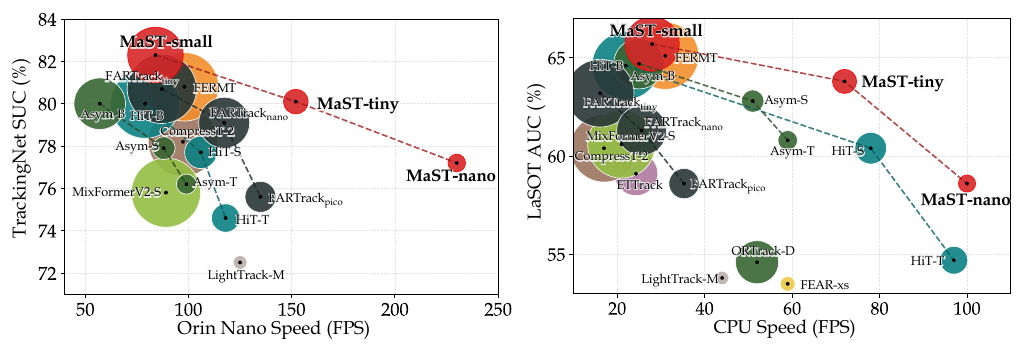}
  \caption{\textbf{Speed–accuracy trade-off on edge platforms.}
  The left panel shows TrackingNet SUC vs.\ Orin Nano speed (FPS).
  The right panel shows LaSOT AUC vs.\ CPU speed (FPS).
  Bubble area is proportional to model parameter count.
  Our MaST-\{small, tiny, nano\} achieve Pareto-optimal trade-offs,
  delivering competitive or superior accuracy at significantly higher
  speeds compared to prior lightweight trackers.
}
  \label{fig:title}
\end{figure*}
Visual object tracking is a fundamental task in computer vision with
broad applications in autonomous driving~\cite{autonomous},
surveillance~\cite{surveillance}, augmented reality~\cite{AR},
and robotics~\cite{robotics}.
Given the target location in an initial frame, the goal is to estimate
its trajectory throughout a video sequence — a task made challenging
by occlusions, background clutter, illumination changes,
and appearance variations.

Recent one-stream Transformer-based trackers~\cite{ostrack,artrack}
have achieved remarkable accuracy by jointly modeling template–search
interactions through self-attention.
However, their quadratic attention complexity over long token sequences
incurs substantial computational cost, hindering deployment on
resource-constrained platforms such as UAVs and mobile robots.
The field has pursued two lines of work to improve efficiency,
including lightweight architectures~\cite{HiT,lighttrack,fear} that employ
scaled-down backbones at the expense of discriminative power,
and adaptive computation methods~\cite{AVTrack,wu2024learning,abavitrack}
that reduce average latency but leave the worst-case computational
cost largely unchanged.

Unlike CNNs that operate on dense feature maps, Vision Transformers
represent images as discrete token sequences, naturally enabling
dynamic token reduction — a strategy proven effective in general
vision tasks~\cite{evit,bolya2022tome}.
This makes token sparsification a particularly promising avenue for
accelerating Transformer-based trackers.
Several recent works have explored this direction.
OSTrack~\cite{ostrack} repurposes the cross-attention scores between
search and template tokens as a pruning criterion, retaining a fixed
ratio of search tokens at intermediate encoder layers;
FARTrack~\cite{wang2026fartrack} extends this by additionally leveraging
attention to learnable location tokens;
AVTrack~\cite{AVTrack} trains an auxiliary sub-network to predict
per-token importance for early dropping decisions.

\begin{figure*}[t]
  \centering
  \includegraphics[width=\linewidth]{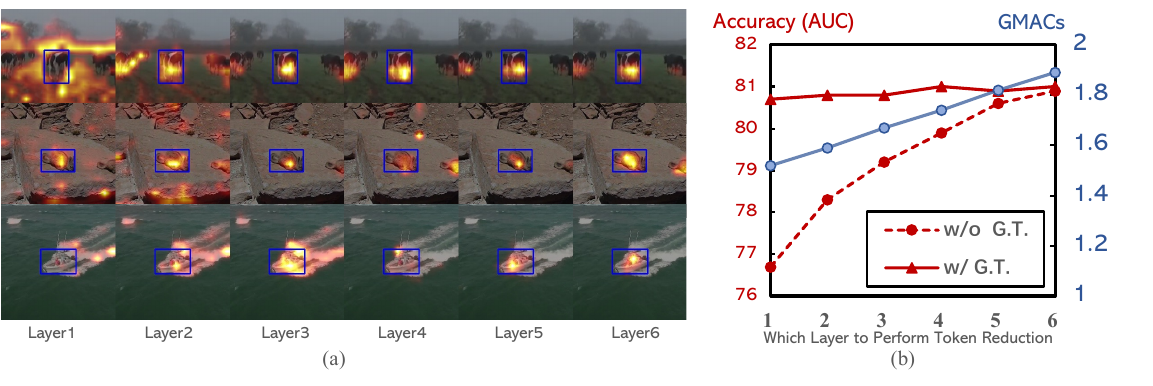}
  \caption{\textbf{Why early-stage token pruning fails.}
  (a)~Attention map visualization across encoder layers on three tracking sequences.
  Early layers produce diffuse, noisy attention that has not yet localized the target,
  explaining why existing methods conservatively defer token pruning to intermediate layers.
  (b)~Effect of performing one-shot token reduction (33\% retention align to OSTrack\cite{ostrack}) at different
  encoder layers on GOT-10k \textit{val}.
  Without spatial guidance (\textit{w/o\,G.T.}), accuracy drops sharply as pruning
  moves to earlier layers.
  With ground-truth target locations guiding selection (\textit{w/\,G.T.}),
  accuracy remains nearly flat across all layers while GMACs decrease
  substantially with earlier pruning.
}
  \label{fig:motivation}
\end{figure*}

Despite these advances, existing token sparsification approaches for
tracking still face systematic limitations. On the token-reduction side,
pruning is typically applied too late, as illustrated in~\figref{fig:motivation}(a),
early-layer attention maps are highly diffuse, making attention-based
importance scores unreliable and forcing most methods to defer pruning
to intermediate layers (e.g., progressively at the 4th, 7th, and 11th layers),
leaving the computationally expensive early layers to process the full
token set. Moreover, common pruning criteria (cross-attention scores or
auxiliary predictors) capture only appearance-level relevance within the
current frame and ignore the strong temporal motion continuity in
tracking, i.e., the prior knowledge of where the target was in the
previous frame. On the output side, the prediction
head remains a dense bottleneck. Convolutional heads assume a dense,
spatially structured token grid, so aggressive sparsification requires
padding and reshaping sparse tokens back to a full 2D feature map,
wasting computation on empty positions; at high compression ratios, the
target-center token may even be pruned, leading to severe
mislocalization. This mismatch between a sparse backbone and a dense
head caps the achievable speedup and degrades accuracy. Notably, the
decoding head is not necessarily convolutional in modern trackers, and some
methods decode targets via additional learnable tokens (e.g.,
MixFormerV2~\cite{mixformerv2} and the ARTrack series~\cite{artrack,wang2026fartrack}),
and LoRAT~\cite{lin2024loratrack} replaces the convolutional head
with MLPs. However, these predictor designs are typically developed
under dense token/feature assumptions, and their behavior under
aggressive token sparsification remains under-explored.

A natural question arises as to whether early-layer pruning is inherently
harmful, or whether it merely lacks the right guidance.
To answer this, we conduct a diagnostic experiment that disentangles
the effect of when to prune from what to prune.
As shown in~\figref{fig:motivation}(b), when token selection at
33\% retention is guided by ground-truth target locations, accuracy
remains remarkably stable regardless of the pruning layer, while computational cost (GMACs) decrease substantially with earlier pruning.
In contrast, without such guidance, accuracy degrades sharply as
pruning moves to earlier layers.
This result indicates that the performance degradation associated with
early pruning stems not from insufficient token representations, but
from the inability to identify the correct tokens to retain at that
stage.
The challenge therefore reduces to two sub-problems,
(i)~obtaining a reliable motion prior for token selection
before the encoder has produced discriminative features, and
(ii)~designing a prediction head that natively operates on the
resulting sparse, unstructured token set.

In this work, we propose \textbf{MaST}
(\textbf{M}otion-\textbf{a}ware \textbf{S}parse \textbf{T}racker),
a sparse tracking framework that addresses these
limitations with a cohesive design.
As shown in~\figref{fig:title}, MaST variants achieve
Pareto-optimal speed–accuracy trade-offs on edge platforms.
MaST performs early, task-driven token reduction in the search region by
injecting a history-based motion prior into the selection process.
To avoid a dense output bottleneck, MaST also introduces a natively
\textbf{sparse prediction head} that operates directly on unstructured
tokens with a score-first, regress-once design, eliminating dense
reshape and reducing redundant computation.

Our main contributions are summarized as follows.
\begin{itemize}
  \item We propose MaST, a sparse tracking framework that performs early,
        task-driven token reduction in the search region using temporal
        motion priors.

  \item We introduce a natively sparse prediction head with a
        score-first, regress-once design that operates directly on
        unstructured tokens, removing the dense head bottleneck.

  \item Extensive experiments on LaSOT, TrackingNet, GOT-10k, and more
        benchmarks demonstrate that MaST achieves competitive or superior
        accuracy while delivering substantially higher measured inference
        speeds on edge platforms such as Raspberry Pi and Jetson Nano.
\end{itemize}

\section{Related Works}
\label{RelatedWorks}
\textbf{Lightweight Visual Tracking.}
Modern one-stream Transformer trackers~\cite{ostrack, mixformer, mixformerv2, artrack, simtrack} unify feature extraction and relation modeling into a single ViT backbone, achieving strong accuracy but suffering from quadratic self-attention complexity that hinders deployment on edge devices.
Prior efforts to close this gap fall into three categories, each with notable limitations.
Compact model design~\cite{fear, HCAT, ETTrack, HiT, LiteTrack, AVTrack} reduces parameter count but sacrifices representational capacity.
Model compression via layer pruning~\cite{LiteTrack, mixformerv2} or knowledge distillation~\cite{compressTracker, mixformerv2} risks non-trivial accuracy degradation.
Conditional dynamic computation~\cite{AVTrack, wu2024learning, abavitrack} improves \emph{average} throughput but yields variable per-frame latency, complicating hardware scheduling.
In contrast, our approach uses fixed top-$K$ token selection guided by a motion prior, achieving both predictable latency and high accuracy.

\textbf{Temporal Priors in Visual Tracking.}
Temporal priors, especially motion locality, are well-established in visual tracking, as target motion between consecutive frames is typically small relative to the search region. Classical approaches exploit this prior via a penalty window (Hanning/cosine) applied to the output score map. SiamFC~\cite{SiamFC} first applied a cosine window to the cross-correlation response to suppress high-response regions near the boundary of the search area. SiamRPN++~\cite{SiamRPN++} further combines a cosine window penalty with a size-change penalty on the regression response to reduce large, abrupt displacement predictions. Discriminative correlation filters such as KCF~\cite{KCF} and ECO~\cite{ECO} apply a cosine window to the input features to reduce periodic boundary artefacts, which also implicitly penalises predictions near the edges. Many modern trackers, including DiMP~\cite{DiMP}, prDiMP~\cite{prDIMP}, STARK~\cite{STARK}, and OSTrack~\cite{ostrack}, likewise multiply a Hanning window onto the predicted score map as a post-processing step before selecting the final target location.
Beyond hand-crafted windows, recent Transformer-based trackers also explore data-driven temporal modeling by encoding historical predictions as tokens and letting the network learn to use this information, e.g., SwinTrack~\cite{lin2022swintrack} and the ARTrack series~\cite{artrack,wang2026fartrack}. However, these designs do not explicitly impose a spatial locality prior during computation.
Despite their widespread adoption, penalty windows are applied only after the full forward pass and merely re-weight already-computed scores, leaving the heavy encoder computation unaffected. In contrast, our approach injects temporal motion priors into the token reduction process itself, using them to guide which tokens are computed in subsequent layers and thereby reducing computation.

\textbf{Token Sparsification in Vision Transformers.}
The tokenized processing paradigm of Vision Transformers enables novel sparsification strategies that are absent in CNN-based trackers. EViT~\cite{evit} pioneered token pruning by selecting tokens with the highest attention to the class token. OSTrack~\cite{ostrack} adapted this idea by measuring cross-attention between search and template tokens, retaining the top-K most correlated tokens. However, their reliance on early-layer attention scores introduces noise. 
As a result, many methods apply token pruning conservatively and progressively at intermediate layers, leading to limited speedup.
DynamicViT~\cite{dynamicvit} proposed learnable importance predictors using lightweight MLPs, and DToP~\cite{DTP4SS} further explored this idea in semantic segmentation. Similarly, GRM~\cite{GRM} applied such techniques to guide token interactions in tracking tasks, while AVTrack~\cite{AVTrack} uses a sub-network for early token exit. SGL~\cite{sgl} likewise observes that partial early-layer attention is insufficient for reliable token pruning and introduces a small vision-language model as external guidance for a larger model. MaST shares the motivation of guided early pruning, but specializes it to single-object tracking by using a nearly cost-free temporal motion prior rather than an auxiliary model. Moreover, our fixed top-$K$ budget avoids the variable per-frame computation of adaptive exits and provides predictable latency.

\begin{figure*}[t]
\begin{center}
\centerline{\includegraphics[width=\textwidth]{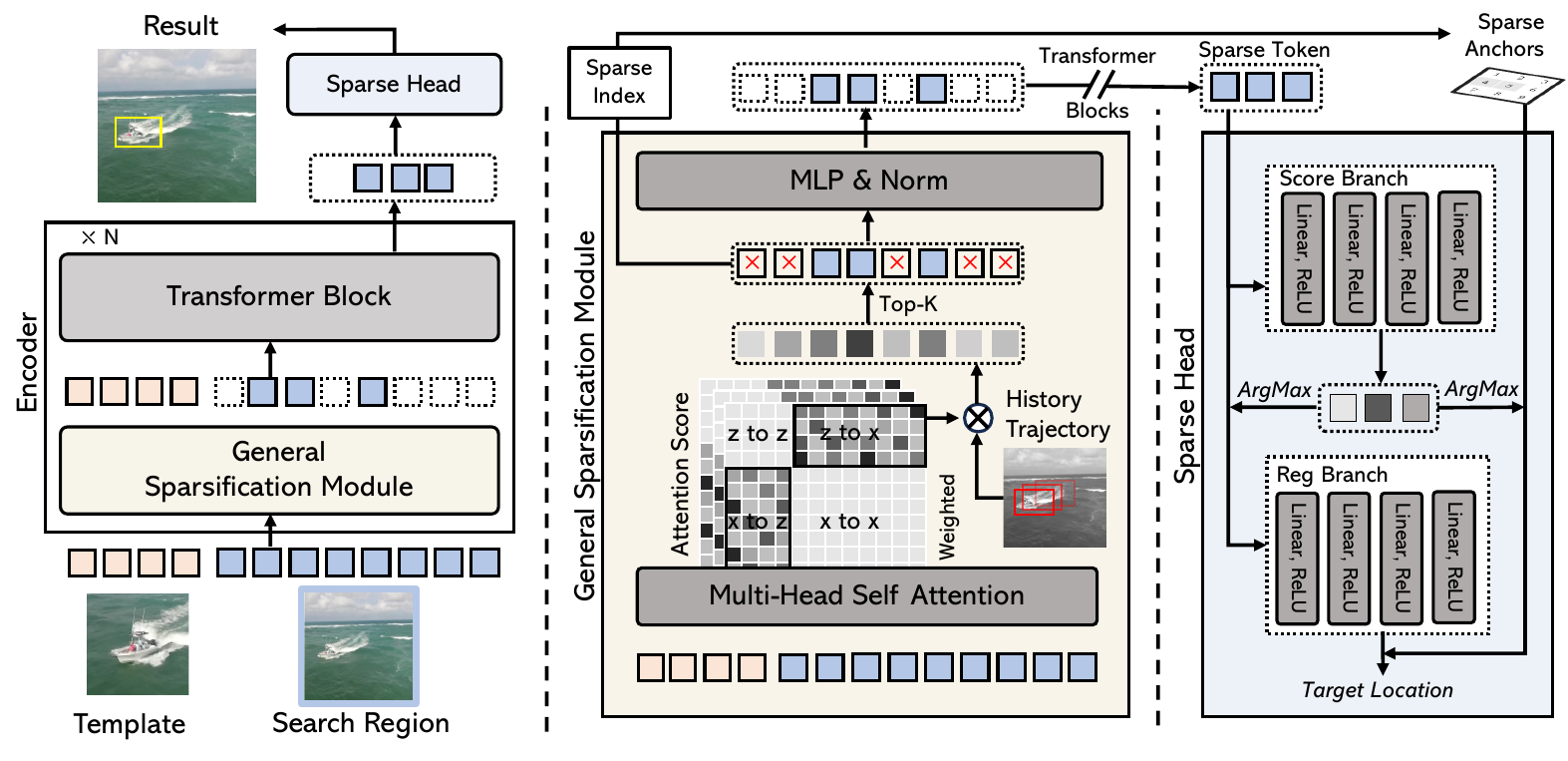}}
\caption{Overview of the MaST framework. The General Sparsification Module is inserted \emph{once} at an early encoder layer, retaining the top-$K$ search tokens; all subsequent Transformer blocks operate on this reduced set. Token selection fuses cross-attention relevance with a motion prior from the previous prediction for early and stable pruning. The retained sparse tokens are then decoded by a lightweight MLP sparse head with a score-first, regress-once pipeline, eliminating dense reshaping and redundant computation.}
\label{main_frame}
\end{center}
\end{figure*}

\section{Method}
\label{Method}

\subsection{Overview}
The proposed framework shown in ~\figref{main_frame} operates in three stages, namely motion prediction, token sparsification, and target decoding. Given an input video frame and the target's initial location, a pair of images is processed as input, consisting of a template image denoted as $Z \in \mathbb{R}^{3 \times H_{z} \times W_{z}}$ and a search image denoted as $X \in \mathbb{R}^{3 \times H_{x} \times W_{x}}$. These images are divided into non-overlapping patches of size $P \times P$, resulting in $P_{z} = \frac{H_{z} \times W_{z}}{P^2}$ patches for the template and $P_{x} = \frac{H_{x} \times W_{x}}{P^2}$ patches for the search region. After patch embedding concatenated together, features are simultaneously extracted and fusion in the form of sequence using a Transformer encoder, in which the sequence will be first handled by our proposed sparsification block. In the end, the retained tokens will be further processed by the prediction head.

\subsection{Motion-Aware Token Sparsification}
Token sparsification plays a critical role in reducing the computational cost of Transformer-based visual object trackers by selectively focusing on the most relevant tokens. Typically, an importance score is assigned to each search token to determine whether it should be retained. Existing approaches often derive these scores either from an auxiliary prediction network or from cross-attention maps that measure the interaction between search and template tokens. Following OSTrack~\cite{ostrack}, we adopt the cross-attention map as the basis for token scoring. However, early-layer attention is often diffuse and noisy, which makes naive early pruning unreliable. To enable earlier and more stable token reduction, we inject a lightweight temporal motion prior derived from the previous prediction.

\textbf{Importance Score.} In our framework, the importance score of search tokens is derived from the cross-attention map in the Transformer encoder. Let the search tokens and template tokens be represented as $\mathbf{Q}_x \in \mathbb{R}^{P_x \times d}$ (queries) and $\mathbf{K}_z \in \mathbb{R}^{P_z \times d}$ (keys), where $P_x$ and $P_z$ are the numbers of tokens in the search and template regions, respectively, and $d$ is the feature dimension. Following the approach explored in OSTrack, we simplify the aggregation process by only using the token from the center patch of the template as a representative feature. This reduces unnecessary complexity while retaining strong performance.

The cross-attention map $\mathbf{A}_{x \to z} \in \mathbb{R}^{P_x \times P_z}$, which measures the interaction between search and template tokens, is defined as:

\begin{align}
\mathbf{A}_{x \to z} = \text{Softmax} \left( \frac{\mathbf{Q}_x \mathbf{K}_z^\top}{\sqrt{d}} \right),
\end{align}

where $\mathbf{A}_{x \to z}[i, j]$ represents the attention weight between the $i$-th search token and the $j$-th template token. Given that we only use the center token of the template, denoted as $\mathbf{k}_c \in \mathbb{R}^d$, the importance score $\mathbf{s}_i$ for the $i$-th search token is simplified as:

\begin{align}
\mathbf{s}_i = \frac{\exp \left( \mathbf{q}_i^\top \mathbf{k}_c / \sqrt{d} \right)}{\sum_{k=1}^{P_x} \exp \left( \mathbf{q}_k^\top \mathbf{k}_c / \sqrt{d} \right)},
\end{align}

where $\mathbf{q}_i \in \mathbb{R}^d$ is the query feature of the $i$-th search token. This formulation focuses the importance scoring process on the interaction between search tokens and the central template token, providing a lightweight yet effective means of measuring token relevance.

\textbf{Motion Prior Injection.} Directly using early-layer cross-attention scores for pruning is sub-optimal due to their diffuse and noisy nature. Tracking, however, provides a strong temporal cue, as the target location changes smoothly across frames. We therefore build a simple spatial prior from the previous prediction and use it to guide token selection.

Specifically, as illustrated in the center panel of~\figref{main_frame}, we construct a Motion Window over the search region, centered at the previous predicted target location, and use it to softly re-weight the cross-attention importance scores. In practice, we find that a 2D Gaussian provides an effective and lightweight realization. Given the predicted bounding box from the previous frame $\mathbf{b}_{t-1} = (x_{t-1}, y_{t-1}, w_{t-1}, h_{t-1})$, where $(x_{t-1}, y_{t-1})$ is the box center and $w_{t-1}$, $h_{t-1}$ are its width and height, the Gaussian is defined as:

\begin{align}
    \mathcal{G}_t(u, v) = \exp \left( -\frac{(u - x_{t-1})^2}{2 \sigma_x^2} - \frac{(v - y_{t-1})^2}{2 \sigma_y^2} \right),
\end{align}

where $(u, v)$ represents the spatial coordinates of a token in the search region, and $\sigma_x = \gamma w_{t-1}$, $\sigma_y = \gamma h_{t-1}$ are the standard deviations of the Gaussian in the $x$ and $y$ directions, respectively. The scaling factor $\gamma$ controls the size of the window, balancing the trade-off between including potentially relevant tokens and excluding irrelevant ones.  We found that setting $\gamma = 0.5$ yields reasonable performance.

The Motion Window assigns higher weights to tokens closer to the previous predicted target center. We refine the importance scores by combining the cross-attention importance scores $\mathbf{s}_i$ with the Motion Window values $\mathcal{G}_t(u_i, v_i)$ for each token:

\begin{align}
\mathbf{w}_i = \mathcal{G}_t(u_i, v_i) \cdot \mathbf{s}_i,
\end{align}

where $(u_i, v_i)$ is the spatial location of the $i$-th token. This combined score $\mathbf{w}_i$ ensures that tokens within the likely target region (as constrained by the Motion Window) are prioritized, even if their raw attention scores are diffuse. Then we only keep the top-$K$ tokens $\mathcal{L}_K\in\mathbb{R} ^{N_K\times d}$ as the output of Motion-Aware Token Sparsification block, attached by the standard Transformer blocks in the encoder.

\subsection{Fully Sparse Prediction Head}
\begin{wrapfigure}{r}{0.5\textwidth}
    \centering
    \includegraphics[width=0.5\textwidth, trim=4pt 5pt 0pt 0pt, clip]{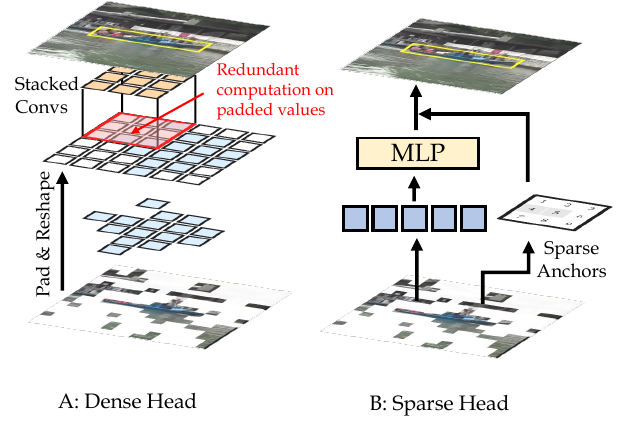}
    \caption{\textbf{Dense vs.\ sparse prediction head.}
    (A)~Dense head pads and reshapes sparse tokens into a full 2D grid, wasting computation on empty positions.
    (B)~Our sparse head applies an MLP directly on retained tokens via sparse anchors, with no padding or reshape.}
    \label{fig:sparse_head}
\end{wrapfigure}
Our goal is to avoid the dense-head bottleneck when the backbone operates on a sparsified, unstructured token set. Let the retained search tokens be
\(\mathcal{L}_K = \{\mathbf{f}_k\}_{k=1}^{N_K}\), where $N_K\ll H_xW_x/P^2$ and each token keeps its original 2D location on the patch grid, denoted as $\mathbf{p}_k=(u_k,v_k)$.
As illustrated in~\figref{fig:sparse_head}, unlike dense heads that must pad and reshape sparse tokens back to a full 2D grid before stacked convolutions, we design a natively sparse head that directly consumes $\mathcal{L}_K$ without padding or reshaping back to a dense feature map.
We parameterize both branches with lightweight MLPs, similar to prior MLP-based heads (e.g., LoRAT~\cite{lin2024loratrack}). We further tailor the head for sparse token decoding via a score-first, regress-once computation path.

\textbf{Score-first, regress-once.} As illustrated in~\figref{main_frame}, the head contains two lightweight branches.
\begin{itemize}
  \item \textbf{Score branch} $g_{\text{s}}$ predicts a scalar confidence $s_k\in\mathbb{R}$ for each retained token.
      The target token is selected by
      \begin{equation}
        k^* = \arg\max_k\; s_k.
      \end{equation}
      This selection can be implemented by scattering $\{s_k\}$ to a sparse score map aligned with the original grid and applying an \texttt{ArgMax}, but no dense feature computation is required.

  \item \textbf{Regression branch} $g_{\text{r}}$ predicts box parameters only \emph{once} for the selected token $\mathbf{f}_{k^*}$.
      Denoting the regression output as $\Delta_{k^*}=(\delta_{x,k^*},\delta_{y,k^*},w_{k^*},h_{k^*})$, the final box is decoded as
      \begin{equation}
        \hat{\mathbf{b}} =
        \mathrm{Decode}(\mathbf{p}_{k^*},\, \Delta_{k^*})
        =
        (u_{k^*}+\delta_{x,k^*},\,v_{k^*}+\delta_{y,k^*},\,w_{k^*},\,h_{k^*}),
      \end{equation}
      followed by conversion from patch-grid coordinates to search-image pixels using the patch stride.
\end{itemize}

This is the same local anchor-based box definition used by dense anchor-based heads, which decode the bounding box at every grid location after a 2D reshape. MaST instead retrieves $\mathbf{p}_{k^*}$ directly from the retained token's pre-pruning flattened-grid coordinate and evaluates the regression branch only at that location. Thus, only the dense reshape and redundant regressions are removed; the box parameterization is unchanged.

This design makes the head complexity scale with $N_K$ for scoring and constant-time for regression, matching the sparsified backbone and eliminating redundant computation on empty spatial positions.

\textbf{Training objective.} We supervise the score branch with a classification loss over sparse tokens, and supervise the regression branch on a single token aligned with the ground-truth target.
Let $\mathbf{c}$ be the ground-truth target center on the patch grid and define
\begin{equation}
  k^{\text{gt}} = \arg\min_k \lVert \mathbf{p}_k - \mathbf{c} \rVert_2.
\end{equation}
The overall head loss is
\begin{equation}
  L_{\text{head}} = L_{\text{cls}}(\{s_k\}) + \lambda_{\ell_1} L_{\ell_1}(\hat{\mathbf{b}}_{k^{\text{gt}}}, \mathbf{b}) + \lambda_{\text{GIoU}} L_{\text{GIoU}}(\hat{\mathbf{b}}_{k^{\text{gt}}}, \mathbf{b}),
\end{equation}
where $\mathbf{b}$ is the ground-truth box, and we set $\lambda_{\text{GIoU}} = 2$ and $\lambda_{\ell_1}=5$.

\section{Experiment}
\label{Experiment}

\begin{wraptable}{r}{0.40\textwidth}
    \centering
    \caption{Details of our MaST model variants.}
    \label{tab:model-card}
    \setlength{\tabcolsep}{3pt}
    \renewcommand{\arraystretch}{1}
    \small
    \resizebox{\linewidth}{!}{
        \begin{tabular}{l | c c c}
            \toprule
            \multirow{2}{*}{Model} & \multicolumn{3}{c}{MaST}                       \\
            \cmidrule(l){2-4}
                                   & nano                     & tiny     & small    \\
            \midrule
            Backbone               & \multicolumn{3}{c}{ViT-Tiny} \\
            Encoder Layers         & 8                        & 12       & 12       \\
            Template Size          & 128                      & 128      & 192      \\
            Search Size            & 256                      & 256      & 384      \\
            Token Rate (\%)        & 30                       & 30       & 30       \\
            MACs (G)               & 0.59                     & 0.84     & 1.82     \\
            Params (M)             & 3.85                     & 5.63     & 5.63     \\
            \bottomrule
        \end{tabular}
    }
\end{wraptable}

\textbf{Device.} 
Our MaST models are implemented using Python 3.10 and PyTorch 2.4.1. Training is conducted on a single NVIDIA RTX 3090 GPU. We evaluate inference speed on three representative edge platforms, namely Raspberry Pi 5, Apple M4, and NVIDIA Jetson Orin Nano. 
To ensure a consistent evaluation environment and better reflect real-world application deployment, we export ONNX files with OpSet 17 and benchmark with ONNXRuntime 1.20.1 to report practical on-device performance. For trackers whose source code is publicly available, we re-evaluate their speed under the same protocol to ensure a fair comparison.

\textbf{Model.} 
To demonstrate our method on the low-cost edge device, we use ViT-Tiny~\cite{vit} as the encoder and the training are initialized with the MAE-lite~\cite{wang2023maelite}.
We provide three model variants with different backbones, encoder depths, and token retention rates, as summarized in~\tabref{tab:model-card}.
MaST-nano uses the first 8 layers of ViT-Tiny for maximum efficiency; MaST-tiny employs the full 12-layer ViT-Tiny; and MaST-small increases the input resolution for higher accuracy. All three default variants retain 30\% of the search tokens. A supplementary ViT-Base scaling study further shows that the speedup persists beyond lightweight backbones while reaching the lower accuracy range of heavier trackers.

\textbf{Training.} 
The training splits of COCO~\cite{coco}, LaSOT~\cite{lasot}, GOT-10k~\cite{got10k} TrackingNet~\cite{trackingnet} and VastTrack~\cite{peng2024vasttrack} are used for training. Common data augmentations including horizontal flip and brightness jittering are used in training. The GPU holds 128 image pairs as the batch size. We train the model with AdamW optimizer~\cite{adamw}, set the weight decay to $10^{-4}$, the initial learning rate for the backbone to $4\times10^{-5}$ and other parameters to $4\times10^{-4}$, respectively. The total training epochs are set to 300 with 60k image pairs per epoch and we decrease the learning rate by a factor of 10 after 240 epochs.  The sparsification rate is gradually warmed-up during training following OSTrack~\cite{ostrack}.  More details of training and hyper-parameters can be found in the supplementary material.

\begin{table*}[t]
  \centering
  \caption{State-of-the-art comparisons of efficient tracking on three large-scale benchmarks, grouped by computational cost. \rankfirst{Red} and \ranksecond{blue} denote the best and second-best results within each group. \Bad{Gray} FPS values indicate speeds below VOT realtime setting (20 FPS).}
  \label{tab-sota-efficient}
  \resizebox{0.998\linewidth}{!}{
    \renewcommand{\arraystretch}{1.2}
    \begin{tabular}{c l| c|ccc c ccc c ccc |l rrr}
      \toprule
       & \multirow{2}*{Method}                          & \multirow{2}*{Pub.} & \multicolumn{3}{c}{LaSOT} &                   & \multicolumn{3}{c}{TrackingNet} &  & \multicolumn{3}{c|}{GOT-10k} & MACs              & \multicolumn{3}{c}{FPS}                                                                                                                     \\
      \cline{4-6}\cline{8-10}\cline{12-14}\cline{16-18}
       &                                                &                     & AUC                       & P$_{Norm}$        & P                               &  & SUC                          & P$_{Norm}$        & P                       &  & AO                & SR$_{0.5}$        & SR$_{0.75}$       & (G)   & RPi5       & CPU        & Nano             \\

      \midrule[0.5pt]
       & KCF~\cite{KCF}                                 & TPAMI'14            & 21.1                      & 22.8              & 18.4                            &  & -                            & -                 & -                       &  & 27.9              & 26.3              & 9.9               & -     & {142}      & {946}      & 203\footnotemark \\
      \rowcolor[gray]{0.92} \multicolumn{18}{l}{\footnotesize\itshape $\sim$500M MACs}                                                                                                                                                                                                                                                                              \\
       & \textbf{MaST-nano}                             & Ours                & \ranksecond{58.6}         & \ranksecond{67.7} & \ranksecond{59.4}               &  & \rankfirst{77.2}             & \rankfirst{82.5}  & \rankfirst{72.4}        &  & 61.6              & \ranksecond{71.8} & \ranksecond{53.7} & 0.585 & {30.1}     & {101}      & {230}            \\
       & AsymTrack-T~\cite{zhu2025asymtrack}            & AAAI'25             & \rankfirst{60.8}          & \rankfirst{68.7}  & \rankfirst{61.2}                &  & \ranksecond{76.2}            & \ranksecond{80.9} & \ranksecond{71.6}       &  & \rankfirst{62.3}  & 71.3              & \rankfirst{54.7}  & 0.708 & \Bad{18.1} & 60         & {99}             \\
       & FEAR-XS~\cite{fear}                            & ECCV'22             & 53.5                      & -                 & 54.5                            &  & -                            & -                 & -                       &  & \ranksecond{61.9} & \rankfirst{72.2}  & -                 & 0.532 & \Bad{19.5} & {59}       & {146}            \\
       & LightTrack~\cite{lighttrack}                   & CVPR'21             & 53.8                      & -                 & 53.7                            &  & 72.5                         & 77.8              & 69.5                    &  & 61.1              & 71.0              & -                 & 0.483 & \Bad{13.4} & 44         & 124              \\
      \rowcolor[gray]{0.92} \multicolumn{18}{l}{\footnotesize\itshape $\sim$1G MACs}                                                                                                                                                                                                                                                                                \\
       & \textbf{MaST-tiny}                             & Ours                & \rankfirst{63.8}          & \rankfirst{72.2}  & \rankfirst{67.2}                &  & \rankfirst{80.1}             & \rankfirst{85.3}  & \rankfirst{77.2}        &  & \rankfirst{66.6}  & \rankfirst{76.0}  & \rankfirst{62.5}  & 0.836 & {22.6}     & {72}       & {152}            \\
       & FARTrack$_\text{pico}$~\cite{wang2026fartrack} & ICLR'26             & 58.6                      & 67.1              & 59.6                            &  & 75.6                         & 81.3              & 70.5                    &  & 62.8              & 72.6              & 50.9              & 1.08  & \Bad{17.2} & 35         & 134              \\
       & AsymTrack-S~\cite{zhu2025asymtrack}            & AAAI'25             & \ranksecond{62.8}         & \ranksecond{71.2} & \ranksecond{64.8}               &  & \ranksecond{77.9}            & \ranksecond{82.2} & \ranksecond{74.0}       &  & \ranksecond{65.5} & \ranksecond{74.8} & \ranksecond{58.9} & 0.806 & \Bad{15.6} & 51         & {88}             \\
       & ORTrack-D~\cite{wu2025ortrack}                 & CVPR'25             & 54.6                      & 62.6              & 54.3                            &  & 73.7                         & 79.1              & 68.2                    &  & -                 & -                 & -                 & 1.54  & \Bad{18.5} & 52         & 138              \\
       & HiT-Small~\cite{HiT}                           & ICCV'23             & 60.5                      & 68.3              & 61.5                            &  & 77.7                         & 81.9              & 73.1                    &  & 62.6              & 71.2              & 54.4              & 1.13  & 21.5       & 78         & {106}            \\
       & HiT-Tiny~\cite{HiT}                            & ICCV'23             & 54.8                      & 60.5              & 52.9                            &  & 74.6                         & 78.1              & 68.8                    &  & 52.6              & 59.3              & 42.7              & 0.995 & {27.3}     & 97         & {118}            \\
       & E.T.Track~\cite{ETTrack}                       & WACV'23             & 59.1                      & -                 & -                               &  & 75.0                         & 80.3              & 70.6                    &  & -                 & -                 & -                 & 1.56  & \Bad{8.8}  & 24         & 71               \\
      \rowcolor[gray]{0.92} \multicolumn{18}{l}{\footnotesize\itshape $\sim$2G MACs}                                                                                                                                                                                                                                                                                \\
       & \textbf{MaST-small}                            & Ours                & \rankfirst{65.8}          & \rankfirst{74.7}  & \rankfirst{70.4}                &  & \rankfirst{82.3}             & \rankfirst{87.1}  & \rankfirst{80.5}        &  & \ranksecond{70.0} & 80.4              & \rankfirst{67.0}  & 1.82  & \Bad{7.5}  & 30         & {98}             \\
       & FARTrack$_\text{nano}$~\cite{wang2026fartrack} & ICLR'26             & 61.3                      & 69.7              & 64.1                            &  & 79.1                         & 84.5              & 75.6                    &  & 69.9              & \rankfirst{81.2}  & 61.4              & 1.78  & \Bad{10.4} & 25         & 117              \\
       & FARTrack$_\text{tiny}$~\cite{wang2026fartrack} & ICLR'26             & 63.2                      & 71.6              & 66.7                            &  & 80.7                         & \ranksecond{85.6} & 77.5                    &  & \rankfirst{70.6}  & \ranksecond{81.0} & \ranksecond{63.8} & 2.65  & \Bad{6.9}  & \Bad{16}   & 87               \\
       & AsymTrack-B~\cite{zhu2025asymtrack}            & AAAI'25             & 64.7                      & 73.0              & 67.8                            &  & 80.0                         & 84.5              & 77.4                    &  & 67.7              & 76.6              & 61.4              & 1.81  & \Bad{6.2}  & 25         & {57}             \\
       & FERMT~\cite{fermt}                             & ECCV'24             & \ranksecond{65.1}         & \ranksecond{74.6} & \ranksecond{69.1}               &  & \ranksecond{80.8}            & 80.9              & \ranksecond{78.1}       &  & 69.6              & 80.1              & 63.2              & 2.31  & \Bad{7.9}  & 31         & {84}             \\
       & HiT-Base~\cite{HiT}                            & ICCV'23             & 64.6                      & 73.3              & 68.1                            &  & 80.0                         & 84.4              & 77.3                    &  & 64.0              & 72.1              & 58.1              & 4.34  & \Bad{6.7}  & 22         & 79               \\
      \bottomrule
      \bottomrule
      \rowcolor[gray]{0.92} \multicolumn{18}{l}{\footnotesize\itshape Heavy Trackers}                                                                                                                                                                                                                                                                               \\
       & SUTrack-T~\cite{sutrack}                       & AAAI'25             & 69.6                      & 79.3              & 75.4                            &  & 82.7                         & 87.2              & 80.8                    &  & 72.7              & 82.1              & 70.5              & 5.52  & \Bad{2.1}  & \Bad{10.9} & 23               \\
       & MCITrack-T~\cite{kang2025mcitrack}             & AAAI'25             & 71.7                      & 81.5              & 78.2                            &  & 84.8                         & 89.4              & 83.7                    &  & 74.0              & 83.9              & 72.1              & 12.6  & \Bad{-}    & \Bad{4.3}  & \Bad{9.1}        \\
       & LoRAT ~\cite{lin2024loratrack}                 & ECCV'24             & 71.7                      & 80.9              & 77.3                            &  & 83.5                         & 87.9              & 82.1                    &  & 72.1              & 81.8              & 70.7              & 19.1  & \Bad{-}    & \Bad{5.4}  & \Bad{12.1}       \\
       & OSTrack~\cite{ostrack}                         & ECCV'22             & 69.1                      & 78.7              & 75.2                            &  & 83.1                         & 87.8              & 82.0                    &  & 71.0              & 80.4              & 68.2              & 18.6  & \Bad{-}    & \Bad{6.5}  & \Bad{13.7}       \\

      \bottomrule
    \end{tabular}
  }

\end{table*}
\footnotetext{KCF is tested on the CPU of Jetson Orin Nano with 256 input size.}
\begin{table*}[t]
  \centering
  \begin{minipage}[t]{0.615\textwidth}
    \centering
\caption{Comparisons with lightweight trackers on more benchmarks.}
\label{tab:more-benchmarks}
\resizebox{\linewidth}{!}{
  \small
  \renewcommand{\arraystretch}{1}
  \begin{tabular}{l | r r | c c c}
    \toprule
    \multirow{2}*{Method}                          & \multicolumn{2}{c|}{FPS} & LaSOT$_\text{ext}$ & UAV123            & NFS                                   \\
                                                   & RPi                      & Nano               & AUC               & AUC               & AUC               \\
    \midrule[0.5pt]
    LightTrack~\cite{lighttrack}                   & 13.4                     & 124                & -                 & 62.5              & 55.3              \\
    FEAR-XS~\cite{fear}                            & 19.5                     & 146                & -                 & 61.4              & 61.4              \\
    E.T.Track~\cite{ETTrack}                       & 8.8                      & 71                 & -                 & 59.0              & 59.0              \\
    HiT-Base~\cite{HiT}                            & 6.7                      & 79                 & 44.1              & 65.6              & 63.6              \\
    MixformerV2-S~\cite{mixformerv2}               & 7.1                      & 89                 & 43.6              & 65.8              & -                 \\
    LiteTrack-B4~\cite{LiteTrack}                  & 3.6                      & 45                 & -                 & 66.4              & 63.4              \\
    CompressTracker-2\cite{compressTracker}        & 6.1                      & 97                 & 40.4              & 62.5              & -                 \\

    AsymTrack-B~\cite{zhu2025asymtrack}            & 6.2                      & 57                 & {44.6}            & \ranksecond{66.5} & 64.4              \\
    FARTrack$_\text{tiny}$~\cite{wang2026fartrack} & 6.9                        & 87                  & \ranksecond{45.0} & 65.8              & \rankfirst{66.9}  \\
    \midrule[0.3pt]
    \textbf{MaST-nano}                             & {30.1}                   & {230}              & 41.1              & 62.9              & 64.4              \\
    \textbf{MaST-tiny}                             & {22.6}                   & {152}              & 42.8              & \rankfirst{66.6}  & \ranksecond{66.2} \\
    \textbf{MaST-small}                            & {7.5}                    & {98}               & \rankfirst{45.3}  & 66.4              & 65.7              \\
    \bottomrule
  \end{tabular}
}

  \end{minipage}
  \hfill
  \begin{minipage}[t]{0.365\textwidth}
    \centering
\caption{Comparison on the VastTrack\cite{peng2024vasttrack} benchmark.}
\resizebox{\linewidth}{!}{
    \begin{tabular}{l | c c c}
        \toprule
        \multirow{2}{*}{Method}                       & \multicolumn{3}{c}{VastTrack}                                        \\
                                                      & SUC                           & P$_{Norm}$        & Prec             \\
        \midrule
        ATOM\cite{ATOM} & 17.2 & 14.1 & 10.2 \\
        SiamRPN++\cite{SiamRPN++}                     & 28.1                          & 29.7              & 24.9             \\
        TransT\cite{TransT}                           & 29.9                          & 31.4              & 25.4             \\
        STARK\cite{STARK}                             & 33.4                          & 34.3              & 30.8             \\
        OSTrack\cite{ostrack}                         & 33.6                          & 34.5              & 31.5             \\
        ARTrack & \rankfirst{35.6} & 35.8 & \ranksecond{32.4} \\
        MixFormerV2-B\cite{mixformerv2}               & \ranksecond{35.2}             & 36.5              & \rankfirst{33.0} \\
        FARTrack$_\text{pico}$\cite{wang2026fartrack} & 30.3                          & 33.0              & 25.7             \\
        FARTrack$_\text{nano}$\cite{wang2026fartrack} & 33.9                          & 35.1              & 30.3             \\
        FARTrack$_\text{tiny}$\cite{wang2026fartrack} & \ranksecond{35.2}             & 36.5              & 32.3             \\
        \midrule[0.3pt]
        \textbf{MaST-nano}                            & 30.6                          & 37.8              & 30.1             \\
        \textbf{MaST-tiny}                            & 33.4                          & \ranksecond{40.7} & 26.2             \\
        \textbf{MaST-small}                           & \rankfirst{35.6}              & \rankfirst{43.2}  & \rankfirst{33.0} \\
        \bottomrule
    \end{tabular}}
\label{tab:vast}

  \end{minipage}
\end{table*}
\subsection{Comparison with State-of-the-arts}
We conduct a comprehensive comparison on five widely used tracking benchmarks. Trackers are grouped into three tiers, namely $\sim$500M, $\sim$1G, and $\sim$2G MACs, based on their computational cost, and compared within the same tier.

\textbf{LaSOT.} LaSOT~\cite{lasot} is a large-scale long-term tracking benchmark comprising 1,400 video sequences with 280 sequences reserved for testing. As shown in~\tabref{tab-sota-efficient}, MaST-tiny achieves the highest AUC of 63.8 and $P_{Norm}$ of 72.2 in the $\sim$1G group, outperforming AsymTrack-S~\cite{zhu2025asymtrack} by +1.0 AUC and HiT-Small~\cite{HiT} by +3.3 AUC, while running at 22.6 FPS on Raspberry Pi 5. MaST-small further pushes to 65.8 AUC in the $\sim$2G group, surpassing FERMT~\cite{fermt} by +0.7 AUC.

\textbf{TrackingNet.} TrackingNet~\cite{trackingnet} is a large-scale short-term benchmark with 511 test sequences covering diverse real-world scenarios. MaST-tiny leads the $\sim$1G group with 80.1 SUC and $P_{Norm}$ of 85.3, surpassing AsymTrack-S by +2.2 SUC. In the $\sim$2G group, MaST-small achieves 82.3 SUC, outperforming all competing methods in its tier.

\textbf{GOT-10k.} GOT-10k~\cite{got10k} is a large-scale dataset with 180 test sequences featuring non-overlapping object categories between train and test splits, designed to measure generalization. MaST-tiny attains an AO of 66.6 and $\text{SR}_{0.75}$ of 62.5 in the $\sim$1G group, outperforming AsymTrack-S by +1.1 AO and +3.6 $\text{SR}_{0.75}$. MaST-small achieves 70.0 AO and the highest $\text{SR}_{0.75}$ of 67.0 in the $\sim$2G group.

\textbf{LaSOT$_{\text{ext}}$.} LaSOT$_{\text{ext}}$~\cite{lasot} extends LaSOT with 150 test sequences covering more diverse and challenging long-term scenarios. As reported in~\tabref{tab:more-benchmarks}, MaST-small achieves the highest AUC of 45.3, surpassing FARTrack$_\text{tiny}$~\cite{wang2026fartrack} by +0.3 and AsymTrack-B~\cite{zhu2025asymtrack} by +0.7. MaST-tiny achieves 42.8 AUC at 22.6 FPS on Raspberry Pi 5, over $3{\times}$ faster than AsymTrack-B (6.2 FPS) ,offering a favorable speed--accuracy trade-off for long-term tracking.

\textbf{UAV123.} UAV123~\cite{uav123} is an aerial tracking benchmark with 123 low-altitude UAV sequences featuring fast motion, small targets, and frequent occlusion. MaST-tiny achieves the highest AUC of 66.6 among all compared methods, outperforming LiteTrack-B4~\cite{LiteTrack} (66.4) and AsymTrack-B~\cite{zhu2025asymtrack} (66.5) while running at 22.6 FPS on Raspberry Pi 5. MaST-nano attains 62.9 AUC at a leading 30.1 FPS, making it well-suited for resource-constrained aerial applications.

\textbf{NFS.} NFS~\cite{nfs} is a high-frame-rate benchmark with 100 sequences featuring fast-moving objects that challenge tracker responsiveness.  Under 30fps setting, MaST-tiny achieves 66.2 AUC, second only to FARTrack$_\text{tiny}$~\cite{wang2026fartrack} (66.9) while running substantially faster on edge devices. MaST-small achieves 65.7 AUC, surpassing AsymTrack-B~\cite{zhu2025asymtrack} (64.4) and HiT-Base~\cite{HiT} (63.6).

\textbf{VastTrack.} VastTrack~\cite{peng2024vasttrack} is a large-vocabulary benchmark spanning 2,115 object categories with over 50,000 sequences, designed to evaluate tracking generalization across a vast range of target types. As shown in~\tabref{tab:vast}, MaST-small achieves a SUC of 35.6 and P$_{Norm}$ of 43.2, surpassing heavier MixFormerV2-B~\cite{mixformerv2}(35.2 SUC) by +0.4. The P$_{Norm}$ advantage is particularly striking at $+6.7$ over the next-best method, suggesting superior localization accuracy normalized by target scale.

\subsection{Ablation Study}

\begin{table}[t]
  \centering
\caption{Comparison of performance on various sparisfication on encoder.
}
\resizebox{0.7\linewidth}{!}{
    \begin{tabular}{l| ccc | c | ccc}
        \toprule
        \multirow{2}*{Sparsification By} & \multicolumn{3}{c|}{{LaSOT}} & \multirow{2}*{\shortstack{Encoder                                    \\ MACs}} & \multicolumn{3}{c}{FPS} \\
                                         & {AUC}                        & P$_{Norm}$                        & P    &       & RPi5 & CPU & Nano \\

        \midrule[0.5pt]
        None                             & 64.0                         & 74.2                              & 68.3 & 1752M & 9.1  & 37  & 94   \\
        Attention                        & 60.5                         & 71.9                              & 65.6 & 824M  & 22.9 & 73  & 157  \\
        Prediction                       & 59.4                         & 71.4                              & 64.9 & 874M  & 21.5 & 66  & 138  \\
        \rowcolor{gray!15}
        Attention + Motion               & 63.8                         & 73.6                              & 67.9 & 824M  & 22.6 & 72  & 152  \\
        Motion                           & 62.6                         & 72.2                              & 66.4 & 824M  & 23.5 & 78  & 169  \\
        \bottomrule
    \end{tabular}}
\label{tab:sparse}

\end{table}
\begin{table}[t]
  \centering
\caption{Comparison of prediction heads on LaSOT with ViT-Tiny backbone.}
\resizebox{0.62\linewidth}{!}{
    \begin{tabular}{l | cc | cc | cc}
        \toprule
        \multirow{2}{*}{Prediction Head}
                                            & \multicolumn{2}{c|}{AUC}
                                            & \multicolumn{2}{c|}{MACs (G)}
                                            & \multicolumn{2}{c}{RPi FPS}                                                     \\
                                            & Dense                         & Sparse & Dense  & Sparse  & Dense  & Sparse \\
        \midrule
        \multicolumn{7}{l}{\textit{Global Localization}} \\
        Loc Tokens~\cite{mixformerv2}       & 59.0                          & 57.9   & 1.76   & 0.845   & 10.0  & 23.9   \\
        Trans. Decoder                      & 61.4                          & 60.1   & 1.88   & 0.833   & 9.4   & 22.7   \\
        \midrule
        \multicolumn{7}{l}{\textit{Anchor-based Localization}} \\
        3$\times$3 Conv~\cite{ostrack}      & 65.0                          & 64.1   & 2.39   & 1.463   & 7.7   & 13.8   \\
        MLP (Dense)~\cite{lin2024loratrack} & 64.0                          & 63.8   & 1.81   & 0.872   & 9.1   & 21.3   \\
        \rowcolor{gray!15}
        MLP (Sparse)                        & {64.0}                        & {63.8} & {1.75} & {0.836} & {9.7} & {23.2} \\
        \bottomrule
    \end{tabular}}
    
\label{tab:head}

\end{table}

\textbf{Impact of Different Sparsification Strategies.}
\tabref{tab:sparse} compares various token sparsification strategies. Naive pruning based solely on attention scores (``Attention'') achieves significant compute reduction but suffers a severe accuracy drop, revealing that early cross-attention scores are too diffuse to reliably identify important tokens. Replacing attention with an MLP predictor (``Prediction'') is even worse, as it lacks spatial grounding. The proposed motion prior is the key enabler, as fusing attention scores with the Gaussian spatial prior (``Attention + Motion'') nearly closes the accuracy gap to the dense baseline while preserving the full compute savings. Using the motion window alone (``Motion'') yields a middle ground, confirming that the fusion of both signals is optimal. These results underscore that a lightweight temporal prior is critical to making aggressive early-layer sparsification practically viable.

\begin{wrapfigure}{l}{0.4\textwidth}
    \centering
    \includegraphics[width=0.3\textwidth]{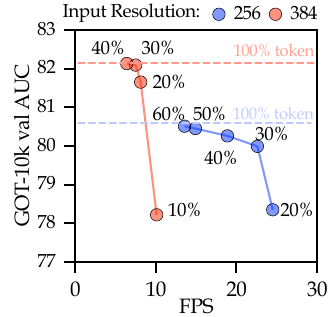}
    \caption{{AUC–FPS trade-off at different token retention rates on GOT-10k val}}
    \label{token_rate}
\end{wrapfigure}

\textbf{Analysis of Different Prediction Heads.}
\tabref{tab:head} compares prediction heads under two paradigms.
\emph{Global localization} heads (Loc Tokens, Trans.\ Decoder) are naturally sparse-compatible and achieve good speedups, but their accuracy is limited (57.9--60.1 AUC) due to insufficient spatial prior. These variants also converge substantially more slowly in image-pair IoU; extending training to $3\times$ the default schedule, initializing from a trained head, and teacher distillation did not close the gap (see the Appendix).
\emph{Anchor-based} heads yield higher accuracy (63.8--64.1 AUC), yet the conventional 3$\times$3 Conv-stacked head must reshape sparse tokens back to a dense grid, incurring redundant computation and capping throughput at 13.8 FPS.
Our sparse MLP head operates score-first, regress-once on retained tokens, improving throughput to 23.2 FPS with no accuracy loss over the dense MLP counterpart~\cite{lin2024loratrack} (63.8 AUC in both cases), confirming the value of a natively sparse head design.

\textbf{Impact of Token Retention Rate.}
\figref{token_rate} examines how the token retention rate affects the AUC–FPS trade-off at two
input resolutions on GOT-10k \textit{val}. For this diagnostic only, we warm up one checkpoint to a 10\% retention rate during training and evaluate the same checkpoint at all plotted rates, rather than retraining a separate model for each operating point.
At 384 resolution, retaining only 30–40\% of tokens already matches the dense baseline in AUC
while substantially boosting throughput; further reduction to 20\% causes only a moderate drop.
At 256 resolution, 50–60\% retention achieves near-lossless accuracy, and 30\% retention remains
competitive.0
\begin{wraptable}{l}{0.4\textwidth}
    \centering
    \caption{Effect of sparsification layer on tracking accuracy and efficiency.}
    \resizebox{\linewidth}{!}{
        \begin{tabular}{c | c c c}
            \toprule
            \shortstack{Sparsify \\ at Layer} & \shortstack{LaSOT \\ AUC} & \shortstack{RPi \\ FPS} & \shortstack{MACs \\ (M)} \\
            \midrule
            \rowcolor{gray!15}
            {1} & {63.82} & {22.6} & {836}  \\
            2          & 63.87          & 17.6          & 942           \\
            3          & 63.65          & 14.0          & 1004          \\
            4          & 63.83          & 12.1          & 1080          \\
            5          & 63.91          & 11.3          & 1154          \\
            6          & 63.96          & 10.1          & 1238          \\
            \bottomrule
        \end{tabular}}
    \label{tab:layer}
\end{wraptable}

In both settings the motion prior guides early token selection reliably regardless of input
scale, confirming that MaST's accuracy–efficiency trade-off is robust across resolution choices and that a single checkpoint supports a smooth post-training speed--accuracy trade-off.

\textbf{Layer-wise Sparsification Analysis.}
\tabref{tab:layer} evaluates the effect of applying token sparsification at different layers (all with motion window) of the ViT-Tiny encoder. Under the motion prior, the choice of sparsification layer has only a negligible impact on accuracy: AUC varies by at most 0.4 points across all configurations. In contrast, the speed penalty of delaying sparsification is substantial, throughput nearly halves from 22.6 to 10.1 FPS when sparsification is pushed to later layers. Sparsifying at Layer 1 therefore offers the best cost-effectiveness, as it matches the accuracy of later-layer alternatives while delivering the highest throughput (22.6 FPS) and the lowest compute (836M FLOPs), and we adopt it as our default.

\textbf{Comparison with Alternative Compression Strategies.}
\tabref{tab:compress} compares MaST against three compression baselines under an identical compute budget ($\approx$1G MACs, ViT-Tiny backbone): the full dense model (1752M MACs), progressive token pruning from OSTrack~\cite{ostrack}, resolution reduction from LoReTrack~\cite{dong2024loretrack}, and layer pruning from CompressTracker~\cite{compressTracker}.
Progressive token pruning retains accuracy (64.0 AUC) but provides negligible speedup (9.1$\to$10.5 FPS on RPi5) due to its multi-stage, non-one-shot design.
Resolution reduction and layer pruning achieve comparable FPS ($\sim$24 FPS on RPi5) but at a severe accuracy cost of $-5.5$ and $-4.9$ AUC, respectively.
Our one-shot token sparsification achieves the lowest MACs (836M) and the best AUC (63.8), within 0.2 of the dense baseline, while maintaining competitive speed (22.6 FPS on RPi5). This demonstrates that spatially-guided one-shot token sparsification offers a clearly superior accuracy--efficiency trade-off compared to resolution reduction or model depth compression.

\begin{table}[ht]
    \centering
    \caption{Comparison with alternative model compression strategies under the same computational budget ($\approx$1G MACs, ViT-Tiny backbone). ``$\downarrow$ Resolution'' reduces the input image size to (48,96); ``$\downarrow$ Layer'' compresses  the encoder into 4 layers; our method applies motion-aware token sparsification at Layer~1 with 30\% retention.}
    \resizebox{0.8\linewidth}{!}{
        \begin{tabular}{l | l | c | c c c | c c c}
            \toprule
            \multirow{2}{*}{Compression Method} & \multirow{2}{*}{From}                  & \multirow{2}{*}{\shortstack{MACs                                                                                           \\(M)}}  & \multicolumn{3}{c|}{LaSOT}            & \multicolumn{3}{c}{FPS}      \\

                                                &                                        &                                  & AUC           & P$_{Norm}$    & P             & RPi5          & CPU        & Nano       \\
            \midrule
            Full model                          & -                                      & 1752                             & 64.0          & 74.2          & 68.3          & 9.1           & 37         & 94         \\
            \midrule
            $\downarrow$ Token (Progressive)    & OSTrack~\cite{ostrack}                 & 1293                             & 64.0          & 74.1          & 68.1          & 10.5          & 41         & 106          \\
            $\downarrow$ Resolution             & LoReTrack~\cite{dong2024loretrack}     & 990                              & 58.5          & 67.7          & 59.4          & 23.6             & 83          & 176          \\
            $\downarrow$ Layer                  & CompressTracker~\cite{compressTracker} & 937                              & 59.1          & 68.5          & 60.1          & 24.1             & 89          & 177          \\

            \rowcolor{gray!15}
            $\downarrow$ Token (One-shot)       & Ours                                   & {836}                     & {63.8} & {73.6} & {67.9} & {22.6} & {76} & {152} \\
            \bottomrule
        \end{tabular}}
    \label{tab:compress}
\end{table}

\begin{figure*}[htbp]
    \begin{center}
    \centerline{\includegraphics[width=0.99\textwidth]{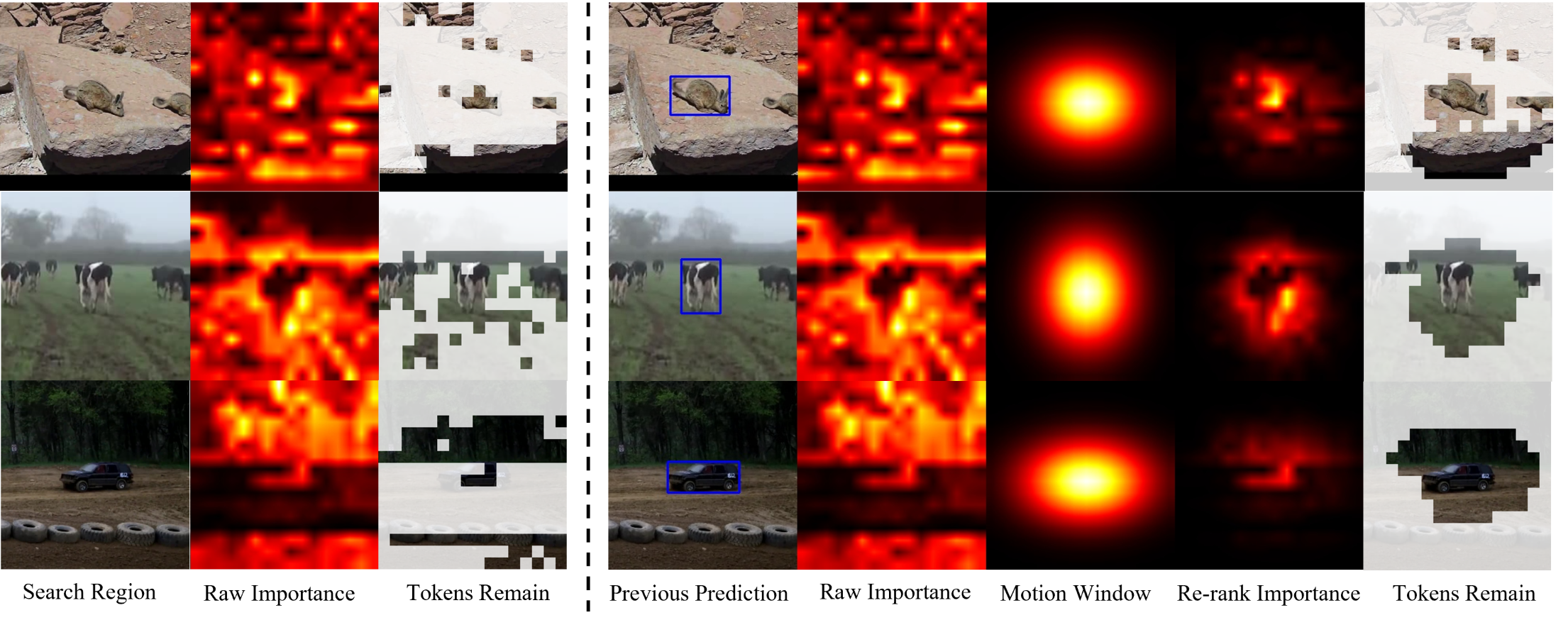}}
    \caption{\textbf{Token sparsification visualization.} \textit{Left} (baseline) shows Search Region, Raw Importance, and Tokens Retained. \textit{Right} (ours) shows Previous Prediction, Raw Importance, Motion Window, Re-ranked Importance, and Tokens Retained. Three tracking sequences are shown per row.}
    \label{vis_window}
    \end{center}
\end{figure*}

\textbf{Visualization of the token sparsification process.}
As shown in~\figref{vis_window}, without a motion prior, raw attention scores are diffuse at early layers, causing retained tokens to scatter across irrelevant background regions. Injecting the motion prior re-ranks scores toward the predicted target location, yielding compact, target-centered retention that directly reduces wasted computation on empty positions.

\section{Conclusion}
In this work, we present MaST, a sparse tracker that unifies motion-aware early token pruning with a natively sparse prediction head. Unlike prior methods that defer pruning to intermediate layers, a motion prior enables reliable one-shot reduction at the first encoder layer, letting all subsequent Transformer blocks operate on the compressed token set. A score-first, regress-once head decodes targets directly from retained tokens, eliminating dense reshape and redundant regression. Together, these designs achieve Pareto-optimal speed--accuracy trade-offs on edge platforms, showing that end-to-end sparsity is a practical path to real-time tracking. A remaining limitation is that high-resolution inputs still incur substantial attention computation before sparsification takes effect, motivating input-adaptive pruning as future work.
\section*{Acknowledgements}
This research is supported by the Major Key Project of Pengcheng Laboratory (PCL2025A13, PCL2025A08), the National Natural Science Foundation of China (U24B20151), and the Guangdong Major Project of Basic and Applied Basic Research (2019B030302002).

\bibliographystyle{splncs04}
\bibliography{main}

\appendix

\section{Appendix}

\subsection{Training Details.}

\textbf{Model Inputs.}
The tracker takes a pair of cropped images as input: a template patch $Z$ and a search region patch $X$, both divided into non-overlapping $16 \times 16$ patches.
Input sizes differ across model variants (see~\tabref{tab:model-card}): MaST-nano and MaST-tiny use a template of $128 \times 128$ and a search region of $256 \times 256$, yielding $64$ template tokens and $256$ search tokens; MaST-small uses a larger template of $192 \times 192$ and a search region of $384 \times 384$, yielding $144$ template tokens and $576$ search tokens.
The backbone is initialized with MAE-lite~\cite{wang2023maelite} pre-trained weights.

\textbf{Augmentation.}
Each training sample is constructed as a template–search image pair sampled from the same video sequence.
Standard data augmentations are applied, including random horizontal flipping and brightness jittering.
The batch size is set to 128 image pairs.

\textbf{Training Pipeline.}
Training proceeds in two stages.
In \textbf{Stage 1}, we train a dense base model (no token sparsification) using the full ViT-Tiny backbone at $128\times128$ template and $256\times256$ search resolution.
We use the AdamW optimizer~\cite{adamw} with a weight decay of $10^{-4}$, a backbone learning rate of $4 \times 10^{-5}$, and $4 \times 10^{-4}$ for all other parameters.
Training runs for 300 epochs with 60{,}000 image pairs per epoch; the learning rate is decayed by a factor of 10 after epoch 240.
This stage produces a single well-converged dense checkpoint shared by all variants.

In \textbf{Stage 2}, we fine-tune three variants from this checkpoint for 50 epochs each with sparsification enabled:
\begin{itemize}
  \item \textbf{MaST-tiny}: fine-tuned directly from the Stage 1 checkpoint at the same $128/256$ resolution.
  \item \textbf{MaST-small}: the position encodings are bilinearly interpolated to accommodate the larger $192\times192$ template and $384\times384$ search resolution, then fine-tuned.
  \item \textbf{MaST-nano}: the last 4 encoder layers are removed and the remaining 8-layer backbone is fine-tuned at $128/256$ resolution.
\end{itemize}

\textbf{Sparsification Warmup.}
Across all Stage 2 variants, token sparsification is introduced with a gradual warmup.
Starting from epoch 1, the token retention rate is linearly annealed from $100\%$ down to the target rate (30\% for all variants) over the first 10 epochs, and then held constant for the remainder of Stage 2 fine-tuning.

\textbf{Motion Prior During Training.}
At test time, the motion window is centered on the bounding box predicted in the previous frame.
During training, the search region is cropped with a random spatial jitter around the ground-truth target location, so the target is approximately centered in the search patch.
We therefore simply set the motion window center to the center of the search image, which naturally approximates the previous-frame prediction without requiring any additional placeholder or annotation.
The motion window is enabled throughout Stage 2 (including during the sparsification warmup epochs).

\begin{table}[ht]
    \centering
    \caption{Comparison of different windowing strategies on GOT-10k \texttt{val}. We evaluate the impact of window type (Gaussian or Cosine), the use of importance scores, and whether the window dynamically adapts to the previous frame’s bounding box size and aspect ratio.}
    \resizebox{0.6\linewidth}{!}{
        \begin{tabular}{c|c|c|ccc}
            \toprule
            {Window} & \multirow{2}{*}{\shortstack{Importance                                   \\Score}} & \multirow{2}{*}{Dynamic} & \multirow{2}{*}{AUC} & \multirow{2}{*}{P} & \multirow{2}{*}{P$_{Norm}$}\\
            {Type}   &                                        &            &      &             \\
            \midrule[0.5pt]
            Gaussian & \checkmark                             & \checkmark & 79.9 & 71.7 & 91.5 \\
            Gaussian & \checkmark                             & -          & 79.5 & 70.5 & 90.2 \\
            Cosine   & \checkmark                             & \checkmark & 79.8 & 71.1 & 91.2 \\
            Cosine   & \checkmark                             & -          & 79.1 & 70.2 & 89.9 \\
            -        & \checkmark                             & -          & 76.9 & 68.2 & 86.5       \\
            Gaussian & -                                      & \checkmark & 78.2 & 69.1 & 88.1      \\
            \bottomrule
        \end{tabular}}
    \label{tab:windowing}
\end{table}
\subsection{Analysis of Motion Window.}
\tabref{tab:windowing} presents an ablation study of various windowing strategies used in the proposed framework. We compare two window types (Gaussian and Cosine), evaluate the effect of incorporating cross-attention importance scores, and analyze whether dynamic window resizing (based on the width and height of the previous frame’s bounding box) improves performance. Results show that combining importance scores with a Gaussian window achieves the best overall performance. The fiexd windowing approach yields sub-optimal perfrmance compared to dynamic window sizes. The last two rows highlight that using only motion windows for sparsification performs better than using only cross-attention scores, but their combination consistently outperforms either individually. This confirms that both the spatial prior and cross-attention scores are crucial for effective token selection.

\textbf{Source of the appearance importance score.}
We additionally compared the center template token used by MaST with max-pooling over all template tokens, following the aggregation considered in GRM~\cite{GRM}. Template max-pooling was slightly less accurate in our experiments and introduced an extra reduction operation with marginal latency. This is consistent with the target-centered construction of tracking templates: the center token is usually a clean foreground cue, whereas pooling over the full template can mix target evidence with background context. We therefore use the center template token as the default appearance-score source.

\begin{table}[ht]
    \centering
    \caption{Results of scaling to a ViT-Base model and increasing input resolution. Accuracy (Acc) and FPS are reported.}
    \resizebox{0.7\linewidth}{!}{
        \begin{tabular}{c|c|c|cc|c}
            \toprule
            Model & Input                    & \%Tokens Retained & LaSOT & GOT-10k & FPS \\
                  &                          &                   & AUC   & AO      & (GPU)   \\
            \midrule[0.5pt]
            \multirow{6}*{\shortstack{ViT-base}}
                  & \multirow{2}*{\(224^2\)} & 100\%             & 71.3  & 71.9    & 116 \\
                  &                          & 30\%              & 69.8  & 70.2    & 215 \\
            \cline{2-6}
                  & \multirow{4}*{\(378^2\)} & 100\%             & 72.1  & 73.3    & 51  \\
                  &                          & 30\%              & 70.4  & 71.5    & 104 \\
                  &                          & 20\%              & 68.9  & 70.6    & 119 \\
                  &                          & 10\%              & 67.2  & 68.4    & 126 \\

            \bottomrule
        \end{tabular}}
    \label{tab:scaling}
\end{table}
\subsection{Scaling to Larger Models}
Experiments shown in~\tabref{tab:scaling} examine the scalability of our motion-aware sparsification by applying it to a ViT-Base tracker (LoRAT~\cite{lin2024loratrack}) and increasing the input resolution from $224^2$ to $378^2$. At $224^2$, 30\% retention nearly doubles throughput from 116 to 215 FPS while attaining 69.8 LaSOT AUC and 70.2 GOT-10k AO. At $378^2$, it more than doubles throughput from 51 to 104 FPS while attaining 70.4 AUC and 71.5 AO. These results place the sparse model in the lower accuracy range of heavier trackers and show that MaST's speedup is not tied to tiny backbones.

\subsection{Additional Details on Global Box Decoding.}
We explored two global localization alternatives that predict box coordinates directly from queries or tokens. \emph{Loc Tokens} appends four box tokens to the encoder, following MixFormerV2~\cite{mixformerv2}; \emph{Trans.\ Decoder} uses four queries that cross-attend to the search tokens after the encoder and then decode box coordinates. Both designs are naturally compatible with sparse tokens, but they converged much more slowly in image-pair IoU. Training for $3\times$ more epochs, replacing the head of a converged tracker, and teacher distillation did not resolve this optimization difficulty. Their LaSOT AUC remained at 57.9--60.1, compared with 63.8 for our local anchor-based sparse head. These attempts motivate preserving the established local anchor definition while removing only its dense implementation.

\begin{table*}[t]
    \centering
    \caption{State-of-the-art comparison on three UAV-oriented tracking benchmarks.
        Methods above the double rule are traditional correlation-filter trackers; those below are deep trackers.
        \rankfirst{Red} and \ranksecond{blue} indicate the best and second-best results overall.}
    \label{tab:uav-sota}
    \resizebox{\linewidth}{!}{
        \renewcommand{\arraystretch}{1.2}
        \begin{tabular}{l |c | c c | c c | c c | c c | c c}
            \toprule
            \multirow{2}*{Method}                     & \multirow{2}*{Pub.} &
            \multicolumn{2}{c|}{DTB70}                &
            \multicolumn{2}{c|}{UAVDT}                &
            \multicolumn{2}{c|}{UAV123}               &
            \multirow{2}*{\shortstack{Param.                                                                                                                                                                                    \\(M)}} &
            \multirow{2}*{\shortstack{MACs                                                                                                                                                                                      \\(G)}} &
            \multicolumn{2}{c}{FPS}                                                                                                                                                                                             \\
            \cline{3-8}\cline{11-12}
                                                      &                     & SUC               & P                 & SUC               & P                & SUC               & P                 &      &       & RPi  & Nano \\
            \midrule[0.5pt]

            HiFT~\cite{HiFT}                          & ICCV'21             & 59.4              & 80.2              & 47.5              & 65.2             & 59.0              & 78.7              & 15.9  & 5.48   & 5.5  & 57   \\
            TCTrack~\cite{TCTrack}                    & CVPR'22             & 62.2              & 81.2              & 53.0              & 72.5             & 60.5              & 80.0              & 9.7  & 8.8   & -    & -    \\
            SGDViT~\cite{SGDViT}                      & ICRA'23             & 60.4              & 78.5              & 48.0              & 65.7             & 57.5              & 75.4              & 23.3 & 11.3  & -    & -    \\
            Aba-ViTrack-DeiT~\cite{abavitrack}             & ICCV'23             & \rankfirst{66.4}  & \ranksecond{85.9} & 59.9              & 83.4             & 66.4              & \ranksecond{86.4} & 7.98  & 2.39   & 7.9    & 83    \\
            SMAT~\cite{SMAT}                          & WACV'24             & 63.8              & 81.9              & 58.7              & 80.8             & 64.6              & 81.8              & 3.76  & 2.49   & 7.7  & 77    \\
            AVTrack-DeiT~\cite{AVTrack}               & ICML'24             & 65.0              & 84.3              & 58.7              & 82.1             & \rankfirst{66.8}  & 84.8              & 7.98  & 2.39   & 7.9 &  83  \\
            ORTrack-DeiT~\cite{wu2025ortrack}         & CVPR'25             & \rankfirst{66.4}  & \rankfirst{86.2}  & \ranksecond{60.1} & 83.4             & 66.4              & 84.3              & 7.98  & 2.39   & 7.9  & 84   \\
            ORTrack-D-DeiT~\cite{wu2025ortrack}       & CVPR'25             & 65.1              & 83.7              & 59.7              & 82.5             & 66.1              & 84.0              & 5.31  & 1.54   & 18.5 & 138  \\
            \midrule[0.5pt]
            \rowcolor[gray]{0.92} \textbf{MaST-nano}  & Ours                & 61.8              & 79.9              & 55.1              & 75.3             & 62.9              & 82.4              & 3.85 & 0.585 & 30.1 & 230  \\
            \rowcolor[gray]{0.92} \textbf{MaST-tiny}  & Ours                & 63.3              & 81.9              & 57.5              & 77.8             & \ranksecond{66.6} & \rankfirst{86.9}  & 5.63 & 0.836 & 22.6 & 152  \\
            \rowcolor[gray]{0.92} \textbf{MaST-small} & Ours                & \ranksecond{66.2} & 85.6              & \rankfirst{63.0}  & \rankfirst{85.9} & 66.4              & 86.2              & 5.63 & 1.82  & 7.5  & 98   \\
            \bottomrule
        \end{tabular}
    }
\end{table*}

\subsection{Comparison on UAV Benchmarks.}
\tabref{tab:uav-sota} provides a comprehensive comparison on three UAV-specific benchmarks.
On UAVDT~\cite{uavdt}, MaST-small achieves the highest success (63.0) and precision (85.9), surpassing ORTrack-DeiT (60.1 / 83.4) by +2.9 / +2.5 while consuming only $1.82$~G MACs versus $2.39$~G; on Jetson Orin Nano, MaST-small also runs faster at 98~FPS versus 84~FPS.
On UAV123~\cite{uav123}, MaST-tiny attains the highest precision of 86.9, outperforming Aba-ViTrack-DeiT (86.4) and AVTrack-DeiT (84.8). It also achieves the second-highest success of 66.6, all with only 0.836~G MACs. On edge devices, MaST-tiny runs at 22.6~FPS on Raspberry Pi~5 and 152~FPS on Jetson Orin Nano.
On {DTB70}~\cite{dtb70}, MaST-small achieves the second-highest success of 66.2 (behind the 66.4 shared by Aba-ViTrack-DeiT and ORTrack-DeiT) and a precision of 85.6, demonstrating competitive accuracy on this challenging aerial benchmark.
MaST-nano is the lightest variant at 0.585~G MACs. It achieves 30.1~FPS on RPi5 and 230~FPS on Jetson Nano, the highest edge throughput among all compared methods, while maintaining solid accuracy across all three benchmarks.
These results confirm that MaST's motion-aware sparse computation transfers favorably to UAV deployment scenarios, delivering a strong accuracy--efficiency trade-off on both general and edge hardware.

\begin{table*}[tb]
  \centering
  \caption{State-of-the-art comparisons of efficient tracking on three large-scale benchmarks, grouped by computational cost. \rankfirst{Red} and \ranksecond{blue} denote the best and second-best results within each group. \Bad{Gray} FPS values indicate speeds below VOT realtime setting (20 FPS).}
  \label{tab-sota-fps}
  \resizebox{0.998\linewidth}{!}{
    \renewcommand{\arraystretch}{1.2}
    \begin{tabular}{c l| c| c c c| c c| c c c c c}
      \toprule
                            & \multirow{2}*{Method}                          & \multirow{2}*{Pub.} & \multirow{2}*{LaSOT} & \multirow{2}*{TrackingNet} & \multirow{2}*{GOT-10k} & \multirow{2}*{\shortstack{Params                                                                   \\(M)}} & \multirow{2}*{\shortstack{MACs\\(G)}} & \multicolumn{5}{c}{FPS} \\
      \cline{9-13}
                            &                                                &                     &                      &                            &                        &                                  &       & RPi5       & M4       & 14900 & Nano             & 3090 \\

      \midrule[0.5pt]
                            & KCF~\cite{KCF}                                 & TPAMI'14            & 21.1                 & -                          & 27.9                   & -                                & -     & {142}      & {946}    & 679   & 203\footnotemark & -    \\
                            & LightTrack~\cite{lighttrack}                   & CVPR'21             & 53.8                 & 72.5                       & 61.1                   & 1.96                             & 0.483 & \Bad{13.4} & 44       & 260   & 124              & 398  \\
                            & FEAR-XS~\cite{fear}                            & ECCV'22             & 53.5                 & -                          & 61.9                   & 6.21                             & 0.532 & \Bad{19.5} & {59}     & 453   & {146}            & 650  \\
      
                            & E.T.Track~\cite{ETTrack}                       & WACV'23             & 59.1                 & 75.0                       & -                      & 6.98                             & 1.56  & \Bad{8.8}  & 24       & 100   & 71               & 194  \\
                            & HiT-Tiny~\cite{HiT}                            & ICCV'23             & 54.8                 & 74.6                       & 52.6                   & 9.59                             & 0.995 & {27.3}     & 97       & 176   & {118}            & 253  \\
                            & HiT-Small~\cite{HiT}                           & ICCV'23             & 60.5                 & 77.7                       & 62.6                   & 11.02                            & 1.13  & 21.5       & 78       & 144   & {106}            & 279  \\
                            & HiT-Base~\cite{HiT}                            & ICCV'23             & 64.6                 & 80.0                       & 64.0                   & 42.14                            & 4.34  & \Bad{6.7}  & 22       & 64    & 79               & 337  \\
                            & FERMT~\cite{fermt}                             & ECCV'24             & 65.1                 & 80.8                       & 69.6                   & 7.98                             & 2.31  & \Bad{7.9}  & 31       & 99    & {84}             & 231  \\
                            & MixformerV2-S~\cite{mixformerv2}               & NeurIPS'23          & 60.6                 & 75.8                       & 61.9                   & 16.05                            & 4.40  & 7.1        & 21       & 72    & 89               & 415  \\
                            & LiteTrack-B4~\cite{LiteTrack}                  & ICRA'24             & 62.5                 & 79.9                       & 65.2                   & 26.48                            & 6.78  & \Bad{3.6}  & \Bad{13} & 36    & 45               & 376  \\
                            & ORTrack-D~\cite{wu2025ortrack}                 & CVPR'25             & 54.6                 & 73.7                       & -                      & 3.76                             & 1.54  & \Bad{18.5} & 52       & 174   & 138              & 480  \\
                            & CompressTracker~\cite{compressTracker}         & ICCV'25             & 60.4                 & 78.2                       & -                      & 21.24                            & 6.38  & \Bad{3.9}  & \Bad{17} & 56    & 97               & 605  \\
                            & AsymTrack-T~\cite{zhu2025asymtrack}            & AAAI'25             & 60.8                 & 76.2                       & 62.3                   & 3.05                             & 0.708 & \Bad{18.1} & 60       & 186   & {99}             & 308  \\
                            & AsymTrack-S~\cite{zhu2025asymtrack}            & AAAI'25             & 62.8                 & 77.9                       & 65.5                   & 3.36                             & 0.806 & \Bad{15.6} & 51       & 165   & {88}             & 269  \\
                            & AsymTrack-B~\cite{zhu2025asymtrack}            & AAAI'25             & 64.7                 & 80.0                       & 67.7                   & 3.36                             & 1.81  & \Bad{6.2}  & 25       & 79    & {57}             & 265  \\

                            & FARTrack$_\text{pico}$~\cite{wang2026fartrack} & ICLR'26             & 58.6                 & 75.6                       & 62.8                   & 2.81                             & 1.08  & 17.2          & 35.3        & -     & 134.9               & -    \\
                            & FARTrack$_\text{nano}$~\cite{wang2026fartrack} & ICLR'26             & 61.3                 & 79.1                       & 69.9                   & 4.59                             & 1.78  & 10.4          & 25.6        & -     & 117.3                & -    \\
                            & FARTrack$_\text{tiny}$~\cite{wang2026fartrack} & ICLR'26             & 63.2                 & 80.7                       & 70.6                   & 6.82                             & 2.65  & 6.9          & 16.1        & -     & 87                & -    \\
      \rowcolor[gray]{0.92} & \textbf{MaST-nano}                             & Ours                & 58.6                 & 77.2                       & 61.6                   & 3.85                             & 0.585 & {30.1}     & {101}    & 247   & {230}            & 556  \\
      \rowcolor[gray]{0.92} & \textbf{MaST-tiny}                             & Ours                & 63.8                 & 80.1                       & 66.6                   & 5.63                             & 0.836 & {22.6}     & {72}     & 184   & {152}            & 453  \\
      \rowcolor[gray]{0.92} & \textbf{MaST-small}                            & Ours                & 65.8                 & 82.3                       & 70.0                   & 5.63                             & 1.82  & \Bad{7.5}  & 30       & 88    & {98}             & 428  \\
      \bottomrule
    \end{tabular}
  }

\end{table*}
\footnotetext{KCF is tested on the CPU of Jetson Orin Nano with 256 input size.}
\subsection{Running Latency on More Platforms.}
Our primary focus is on real-world deployment, particularly on edge devices where power consumption and thermal constraints are critical. High-power GPUs and desktop CPUs are convenient for offline benchmarking but rarely represent embedded or mobile deployment conditions.
Critically, \emph{high throughput on powerful hardware does not imply efficiency on low-power devices}.
Server-grade GPUs and desktop CPUs have large L2/L3 caches and abundant on-chip SRAM. These resources hide irregular memory-access latency and keep arithmetic units saturated. Edge platforms such as Raspberry Pi 5 and Jetson Orin Nano have no such buffer. Their small caches and limited memory bandwidth penalize operations that are cache-friendly on high-end hardware but bandwidth-bound on constrained devices.
This disparity is evident in~\tabref{tab-sota-fps}. CompressTracker reaches 605~FPS on an RTX~3090 but only 3.9~FPS on Raspberry Pi~5, well below the VOT realtime threshold. LiteTrack-B4 similarly reaches 376~FPS on GPU but drops to 3.6~FPS on-device. MaST avoids dense operations that scale poorly on cache-limited hardware. MaST-nano and MaST-tiny sustain 30.1 and 22.6~FPS on Raspberry Pi~5 respectively, while remaining competitive or superior in accuracy.

\subsection{NFS under Different Input Frame Rates.}
\tabref{tab:nfs-framerate} compares MaST-tiny with OSTrack-Tiny under the 30~FPS and 240~FPS evaluation protocols on NFS. At 30~\textit{fps}, MaST-tiny already outperforms OSTrack-Tiny by +1.1 AUC, +0.6 P$_{Norm}$, and +0.3 P. At 240~\textit{fps}, the margin widens to +2.9 AUC, +4.0 P$_{Norm}$, and +4.0 P. Higher frame rates yield smaller inter-frame displacement and a clearer motion prior, allowing MaST to guide token selection more accurately than appearance-only methods.

The comparison also highlights the importance of tracker latency. From 30 to 240~\textit{fps}, OSTrack-Tiny improves only marginally (+0.7 AUC), while MaST-tiny gains +2.5 AUC. A motion-aware tracker can better exploit improved temporal continuity when its inference latency is low. MaST-tiny runs at 152~FPS on Jetson Nano versus 83~FPS for OSTrack-Tiny, making high-frame-rate online tracking more practical in real deployments.

\begin{table}[ht]
    \centering
    \caption{Comparison on NFS\cite{nfs} under different input frame-rate settings. We compare MaST-tiny with OSTrack-Tiny under the standard 30 FPS and 240 FPS protocols. Higher input frame rates provide a more stable motion prior, while lower tracker latency improves the ability to exploit this temporal continuity in practice. Please fill in the numeric entries.}
    \resizebox{0.7\linewidth}{!}{
        \begin{tabular}{l|ccc|ccc|c}
            \toprule
            \multirow{2}{*}{Method} & \multicolumn{3}{c|}{@30 \textit{fps}} & \multicolumn{3}{c|}{@240 \textit{fps}} & Nano \\
            & AUC & P$_{Norm}$ & P & AUC & P$_{Norm}$ & P & FPS \\
            \midrule[0.5pt]
            OSTrack-Tiny~\cite{ostrack} &65.1 & 81.2 & 78.4 & 65.8 & 81.3 & 78.5 & 83 \\
            \textbf{MaST-tiny}          & 66.2 & 81.8 & 78.7 & 68.7 & 85.3 & 82.5 & 152 \\
            \bottomrule
        \end{tabular}}
    \label{tab:nfs-framerate}
\end{table}

\subsection{Robustness to Occlusion and Abrupt Motion.}
MaST does not rely on the motion prior alone. As shown by the main-paper sparsification ablation, fusing appearance attention with motion reaches 63.8 LaSOT AUC, compared with 62.6 for motion alone and 60.5 for attention alone. The Gaussian prior is a soft re-weighting, so strong off-center appearance evidence can still retain tokens outside the most likely motion region.

\begin{figure}[t]
\centering
\includegraphics[width=\linewidth]{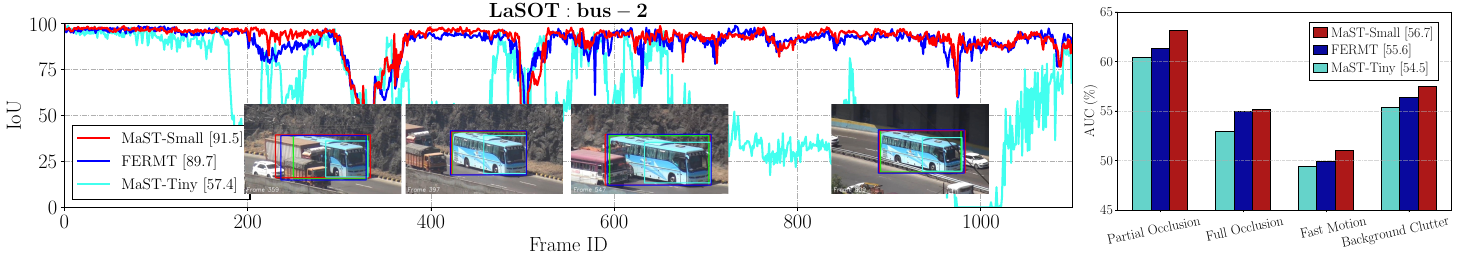}
\caption{\textbf{Robustness analyses.} Left: per-frame IoU on the LaSOT \texttt{bus-2} sequence, with representative predictions around an occlusion. Right: AUC on challenging LaSOT attributes. Under the same $4\times$ search factor and 256-pixel search input as FERMT, MaST-tiny is less robust to severe occlusion and fast motion; increasing the candidate field to a $5\times$ factor and 384-pixel input allows MaST-small to surpass FERMT while using much less computation.}
\label{fig:failure-and-attr}
\end{figure}

\figref{fig:failure-and-attr} illustrates the view--token-budget trade-off. After the bus is occluded, MaST-tiny may retain mainly the visible front of the vehicle and localize only its head. A larger search view gives MaST-small a wider candidate field and improves the challenging-attribute results. This suggests that long occlusion and large displacement are chiefly constrained by the local search crop and its token budget, rather than by a hard dependence on the motion center.

\begin{figure}[t]
\centering
\includegraphics[width=\linewidth]{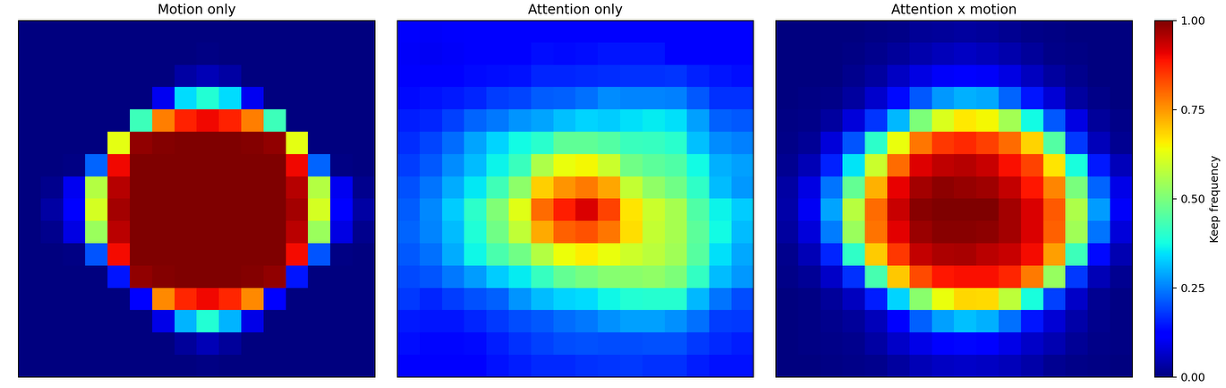}
\caption{\textbf{Token keep frequency on GOT-10k \textit{val}.} We align search crops to a common grid and accumulate binary keep/drop decisions. Warmer cells are retained more often. The product of appearance attention and the soft motion prior concentrates computation near the likely target while retaining off-center evidence.}
\label{fig:keep-frequency}
\end{figure}

\subsection{More Visualizations.}
\begin{figure}[h]
\begin{center}
\centerline{\includegraphics[width=0.98\columnwidth]{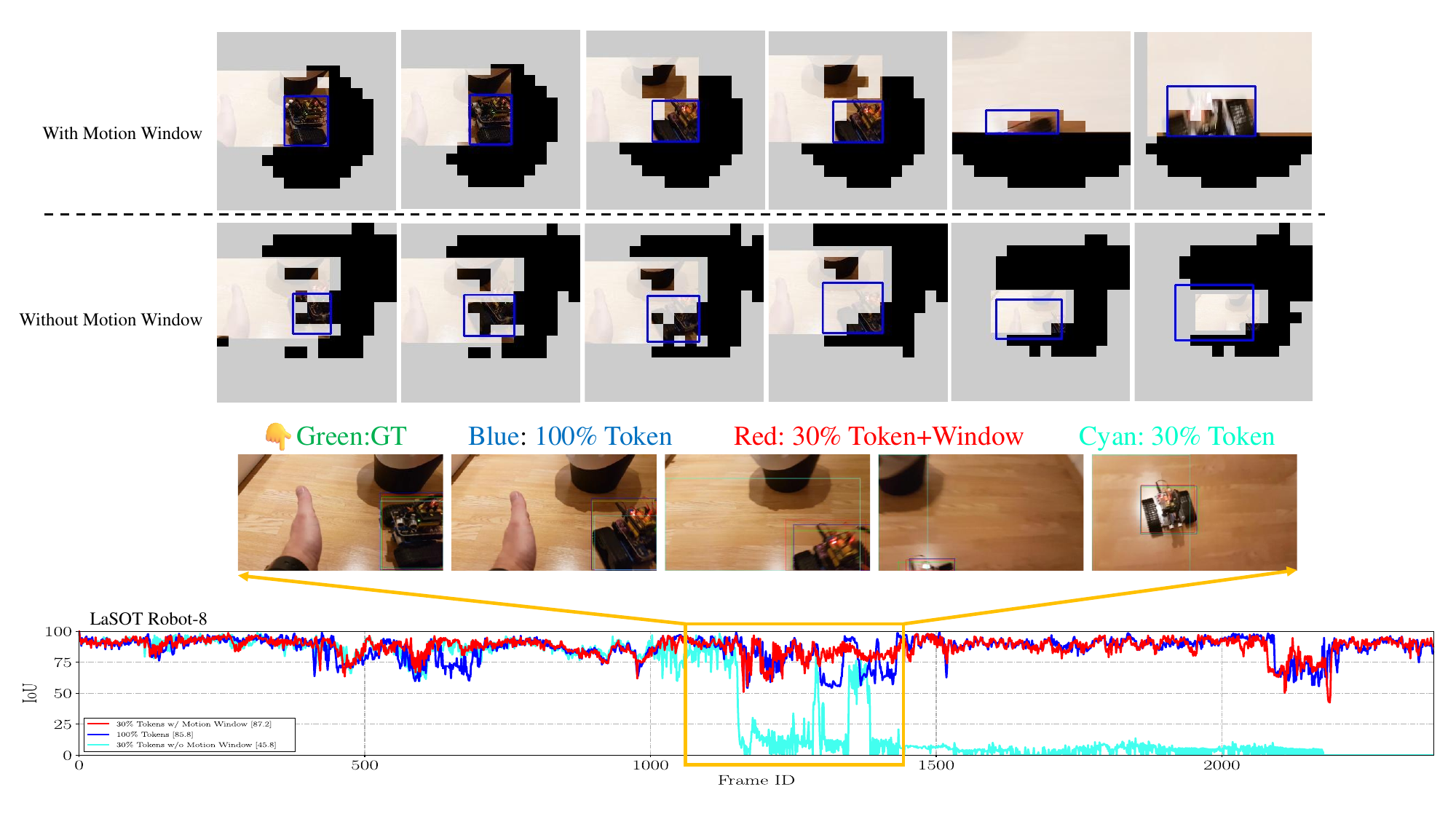}}
\caption{Tracking a robot when it engage fast movement, leading to partially out-of-view.  Sparsification without motion-aware fixing would quickly discard the features of the target, making the tracker focus on other objects in the view.}
\label{fig:vis1}
\end{center}
\end{figure}

1. Rapid Motion Scenario. The visualization of \figref{fig:vis1} demonstrates that MaST (30\% tokens with motion window) can robustly track rapidly moving objects, closely matching the performance of the full-token model (100\% tokens). In contrast, a naive sparsification strategy without the motion-aware component fails completely. This clearly illustrates the crucial role of our motion-aware sparsification in handling rapid motion cases.

\begin{figure}[h]
\begin{center}
\centerline{\includegraphics[width=0.98\columnwidth]{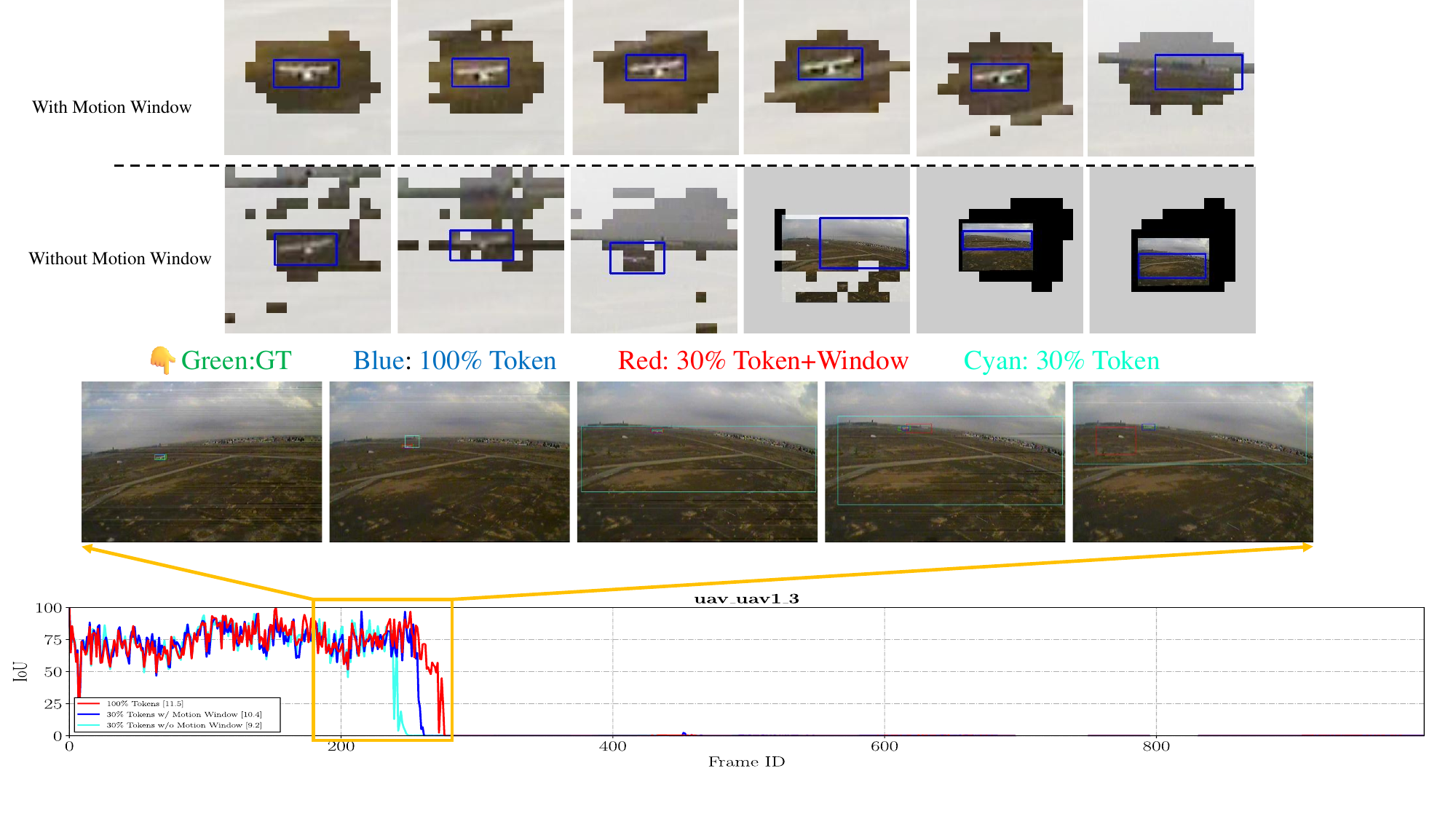}}
\caption{Handling small objects like UAVs under extremely bad situation.  With or without sparsification or motion-aware sparsification, the tracker is difficult to keep tracking the target.}
\label{fig:vis2}
\end{center}
\end{figure}

2. Failure Cases under Extremely Challenging Conditions. We acknowledge there exist extreme conditions, e.g., low-quality UAV footage with significant appearance ambiguity in \figref{fig:vis2}, that surpass the capability of even the full-token model. Such difficulties are fundamentally challenging for existing state-of-the-art trackers, not limited to MaST. Therefore, performance degradation under these conditions stems from inherent limitations common to the tracking domain rather than from sparsification or motion-aware strategies.

\subsection{Limitations.}
MaST inherits the local search region paradigm that is standard across modern one-stream and two-stream trackers.
This design choice means that target loss under fast motion or prolonged occlusion remains an inherent challenge shared with the broader tracking community, rather than a problem introduced by our motion-aware sparsification.
When the inter-frame displacement exceeds the search region radius, or when the target is occluded for an extended period, the tracker cannot relocate the target beyond the local window. This holds regardless of how tokens are selected within it.
Looking forward, we see motion-aware sparse computation itself as a promising tool to address this limitation: by concentrating computation on a small set of informative tokens, the saved budget could be redistributed to extend the effective search area or to maintain a lightweight global context stream, enabling wider spatial coverage without increasing the overall computational cost.
The retention rate is currently fixed for predictable latency. Our rate-sweep experiment shows mild adaptability from a single checkpoint; a stronger policy could vary the rate per frame using target confidence, motion uncertainty, or a device-level compute budget.

\clearpage

\end{document}